%% file: SeeBigFromSmall_camera_ready.tex
\pdfoutput=1

\documentclass[11pt]{article}

\usepackage[preprint]{acl}

\usepackage{times}
\usepackage{latexsym}

\usepackage[T1]{fontenc}

\usepackage[utf8]{inputenc}

\usepackage{microtype}

\usepackage{inconsolata}

\usepackage{graphicx}

\usepackage{booktabs}
\usepackage{array}
\usepackage{multirow}
\usepackage{amsfonts,amssymb}
\usepackage{bm}
\usepackage{amsmath}
\usepackage{microtype}
\usepackage{bm}
\usepackage{pifont}
\usepackage{makecell}
\usepackage{graphicx}
\usepackage{booktabs}
\usepackage{subcaption}
\usepackage{caption}
\usepackage{float} 
\usepackage{adjustbox}
\usepackage{longtable}
\usepackage{amsmath}
\newcommand{\cmark}{\ding{51}}%
\newcommand{\xmark}{\ding{55}}%
\usepackage{fontawesome5}
\usepackage{color}
\usepackage{hyperref}
\title{``Seeing the Big through the Small'': Can LLMs Approximate\\ 
Human Judgment Distributions on NLI from a Few Explanations?}

\author{
 \textbf{Beiduo Chen\textsuperscript{\faMountain\kern1pt\faRobot}} \quad
 \textbf{Xinpeng Wang\textsuperscript{\faMountain\kern1pt\faRobot}} \quad
 \textbf{Siyao Peng\textsuperscript{\faMountain\kern1pt\faRobot}} \quad
 \textbf{Robert Litschko\textsuperscript{\faMountain\kern1pt\faRobot}} \quad
\\
 \textbf{Anna Korhonen\textsuperscript{\faSchool}} \quad
 \textbf{Barbara Plank\textsuperscript{\faMountain\kern1pt\faRobot}}
\\
\textsuperscript{\faMountain} MaiNLP, Center for Information and Language Processing, LMU Munich, Germany \\
\textsuperscript{\faRobot} Munich Center for Machine Learning (MCML), Munich, Germany \\
\textsuperscript{\faSchool} Language Technology Lab, University of Cambridge, United Kingdom \\
{\tt \{\href{mailto:beiduo.chen@lmu.de}{\textcolor{black}{beiduo.chen}}, \href{mailto:xinpeng.wang@lmu.de}{\textcolor{black}{xinpeng.wang}}, \href{mailto:siyao.peng@lmu.de}{\textcolor{black}{siyao.peng}}, \href{mailto:robert.litschko@lmu.de}{\textcolor{black}{robert.litschko}}\}@lmu.de}
\\
\tt 
\href{mailto:alk23@cam.ac.uk}{\textcolor{black}{alk23@cam.ac.uk}},
\href{mailto:b.plank@lmu.de}{\textcolor{black}{b.plank@lmu.de}}
}

\begin{document}
\maketitle

\begin{abstract}

Human label variation (HLV) is a valuable source of information that arises when multiple human annotators provide different labels for valid reasons.
In Natural Language Inference (NLI) earlier approaches to capturing HLV involve either collecting annotations from many crowd workers to represent human judgment distribution (HJD) or use expert linguists to provide detailed explanations for their chosen labels. 
While the former method provides denser HJD information, obtaining it is resource-intensive.
In contrast, the latter offers richer textual information but it is challenging to scale up to many human judges. 
Besides, large language models (LLMs) are increasingly used as evaluators (``LLM judges'') but with mixed results, and few works aim to study HJDs.
This study proposes to exploit LLMs to approximate HJDs using a small number of expert labels and explanations. 
Our experiments show that a few explanations significantly improve LLMs' ability to approximate HJDs with and without explicit labels, thereby providing a solution to scale up annotations for HJD.  
However, fine-tuning smaller soft-label aware models with the LLM-generated model judgment distributions (MJDs) presents partially inconsistent results: while similar in distance, their resulting fine-tuned models and visualized distributions differ substantially. 
We show the importance of complementing instance-level distance measures with a global-level shape metric and visualization to evaluate MJDs more effectively against human judgment distributions.

\end{abstract}

\input{Sections/introduction}
\input{Sections/related_work}

\input{Sections/method}

\input{Sections/experiment_setup}
\input{Sections/results}

\section{Conclusion}
This paper analyzes to what extent LLMs can approximate human judgment distributions from a few explanations. 
Our results show that a few explanations improve LLM's ability to approximate HJDs. 
However, measuring the distance of the resulting MJD is insufficient: while similar in distance, their resulting fine-tuned models and visualized distributions differ substantially. 
We adopt an error visualization tool and a global-level metric, aligning our distribution and fine-tuning results. 
Our method can also be extended to other tasks beyond NLI, and we encourage an uptake of explanation-informed datasets.

\section*{Limitations}
Approximating human judgment distributions from a few explanations is a challenging task.
The generalizability of LLMs empowers our approach to transform textual data (prompts with comments) into numeric form (MJDs) and thus approximate the human label distribution.
However,
we are constrained to existing crowd-annotated NLI datasets as our approximation target, which is not necessarily the best or most representative human label distribution. In future it would be interesting to test our approach on diverse HJDs.  
For the same reason, our fine-tuning experiments use the standard soft-label training, which is one of the most prominent but not the only HLV-embracing procedure. 

There are further considerations we could take into account in our experimental design. 
Firstly, the VariErr dataset also includes a second round of validity judgments regarding the explanations to identify erroneous labels. In the current study, this was not in scope.  
We could further leverage these quality judgments as features to subset better and worse explanations and investigate LLMs' performances when fed with different explanation qualities. 
Secondly, we could further explore the temperature $\tau$ or other normalization ways to scale or zoom in to better understand the differences in shapes between HJDs and MJDs. 
Thirdly, traditional divergence metrics, such as KL, JSD, and TVD, are not particularly designed to measure global-level differences between two distributions, especially concerning the distribution shapes. While we proposed one distance correlation measure to address this, there can be other metrics to measure these macroscopic differences. 
Similarly, error visualizations partially prove our hypothesis but alone do not provide quantifiable evidence. 

\section*{Acknowledgements}
We thank the members of the MaiNLP lab for their insightful feedback on earlier drafts of this paper. 
We specifically appreciate the suggestions of
Philipp Mondorf, Bolei Ma, Kassem Sabeh, Verena Blaschke, Diego Frassinelli, Sondre Wold, Jian Lan, and Rob van der Goot. 
BC acknowledges his membership in the European Laboratory for Learning and Intelligent Systems (ELLIS) PhD program.
We are also grateful to the anonymous reviewers for their constructive feedback. 
This research is supported by ERC Consolidator Grant DIALECT 101043235, and UK Research and Innovation (UKRI) Frontier Research Grant EP/Y031350/1.

\newpage

\appendix

\input{Sections/appendix}

\end{document}

%% file: Sections/introduction.tex
\section{Introduction}

\begin{figure}[t]
\centering
\includegraphics[width=0.48\textwidth]{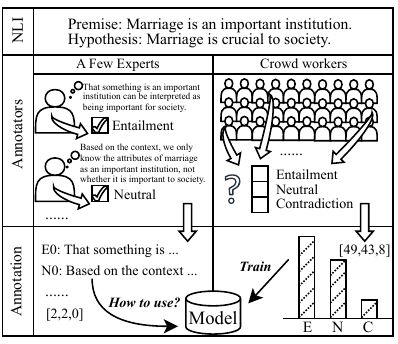}
\caption{Comparison between approaches to investigate HLV in NLI.
Experts first explain the sample individually and then select a label, while crowd workers only record their choices. Explanations provide details for labels to understand HLV.
However, it is not clear how to use explanations effectively to model HLV.}
\label{fig:example}
\end{figure}

\begin{figure*}[t]
\centering
\includegraphics[width=\textwidth]{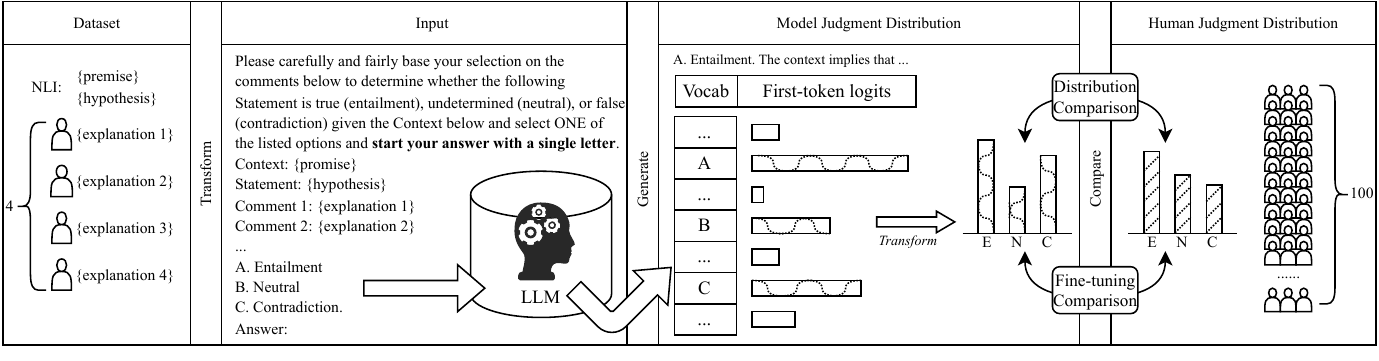}
\caption{The overall structure of our LLM approximation system. 
Explanations from 4 annotators in VariErr NLI \cite{DBLP:journals/corr/abs-2403-01931} are transformed with corresponding NLI samples together into multiple-choice questions, and the generated soft labels (model judgment distributions) are compared with human judgment distributions from 100 crowed workers in Chaos NLI \cite{DBLP:conf/emnlp/NieZB20}.
Two SOTA open-source LLMs, Mixtral \cite{DBLP:journals/corr/abs-2401-04088}, and Llama3 \cite{llama3}, interpret the explanations, and we conduct comparisons on \emph{distribution} and \emph{fine-tuning}.}
\label{fig:overall}
\end{figure*}

In Natural Language Processing (NLP), we are faced with many situations in which more than one plausible label (or reading) exists, a phenomenon referred to as Human Label Variation (HLV,~\citealt{DBLP:conf/emnlp/Plank22}). HLV could be caused by inherent disagreement \cite{DBLP:journals/tacl/PavlickK19}, subjectivity \cite{cabitza2023toward}, or cases where multi-answers are plausible. An increasing body of work suggests that HLV provides rich information that should not be discarded as noise~\cite[e.g.][]{DBLP:journals/aim/AroyoW15,plank-etal-2014-learning, DBLP:journals/jair/UmaFHPPP21} as it impacts every step of the learning process 
\cite{DBLP:conf/emnlp/Plank22}.
As one of the fundamental natural language understanding tasks, Natural Language Inference (NLI) \cite{DBLP:conf/mlcw/DaganGM05,DBLP:conf/emnlp/BowmanAPM15,DBLP:conf/naacl/WilliamsNB18, Manning2006LOCALTI} has embraced HLV especially~\cite[e.g.][]{DBLP:journals/tacl/PavlickK19,DBLP:conf/emnlp/NieZB20,jiang-marneffe-2022-investigating,zhou-etal-2022-distributed}.

There are two common approaches to investigating HLV in NLI, illustrated in Figure~\ref{fig:example}.
One way is to collect annotations from a ``big'' number of crowd workers \cite{DBLP:journals/tacl/PavlickK19}, enabling smoother label probabilities from a statistical standpoint to represent ``collective'' human opinion \cite{DBLP:conf/emnlp/NieZB20}. The obtained human judgment distributions (HJDs) are typically soft labels suitable
for model training and evaluation \cite{DBLP:conf/naacl/FornaciariUPPHP21,DBLP:journals/jair/UmaFHPPP21,DBLP:journals/corr/abs-2403-04085}. 
It provides a rich resource to study and model HLV via soft labels or other methods~\cite{DBLP:journals/jair/UmaFHPPP21,gruber-etal-2024-labels,DBLP:journals/tacl/DavaniDP22,DBLP:conf/naacl/FornaciariUPPHP21,jiang-marneffe-2022-investigating}.
However, this annotation scheme is labor intensive, and despite offering limited insights on how label variation arises
\cite{DBLP:journals/tacl/JiangM22}. 
Contrary to the former dense HJD-based annotation, some NLI variation datasets are annotated by a ``small'' number of expert linguists, delivering annotated labels along with corresponding explanations shown to contain richer linguistic information than explicit labels alone \cite{DBLP:conf/emnlp/JiangTM23,DBLP:journals/corr/abs-2403-01931}, as each annotator explains the label they selected. 
These accompanying explanations
however, 
are scarce, and it is not clear how to use them effectively for modeling. 

LLMs, benefiting from enormous training data, are capable of generalizing across various tasks \cite{DBLP:journals/corr/abs-2303-18223}, 
from text generation \cite{DBLP:conf/emnlp/LeePSWJ23}, model distillation \cite{DBLP:journals/corr/abs-2402-13116}, to name a few, to more recently, functioning as ``LLM judges,'' for example in evaluation~\cite{chiang-lee-2023-large,verga2024replacing} or linguistic annotation~\citep[e.g.][]{ettinger-etal-2023-expert}. 
This suggests that LLMs can serve as a good
bridge between humans and machines.
In this paper, our main research questions are: \textit{Can LLMs provided with a ``small'' number of detailed explanations
better approximate the human judgment distributions collected by a ``big'' number of annotators?} If this is the case, \textit{are the obtained model judgment distributions (MJDs) suitable as soft labels for fine-tuning smaller models to predict distributions?} 
To investigate these questions, we provide two sets of main experiments, as illustrated in 
Figure~\ref{fig:overall}.

Experiments on \textit{distribution comparison} show that a few explanations can improve the capabilities of LLMs to approximate human judgment distributions among various metrics.
To demonstrate the practical significance of the generated MJDs from our method, we applied them in a \textit{fine-tuning comparison} to train
pre-trained transformer models such as BERT \cite{DBLP:conf/naacl/DevlinCLT19} and RoBERTa \cite{DBLP:journals/corr/abs-1907-11692}.
Interestingly, our analysis shows that the effectiveness of the generated MJDs serving as fine-tuning data cannot be predicted well by \textit{distribution comparison}.
We hypothesize that instance-level measures, e.g., KL Divergence \cite{kullback1951information} and Jensen-Shannon Distance \cite{DBLP:journals/tit/EndresS03}, overlook the dependencies across sample points, which captures the global-level HLV useful for training models.
Therefore, we utilize distance correlation \cite{szekely2007measuring} to measure the global-level association between the generated MJDs over the whole dataset and the corresponding target HJD. 
We empirically show that distance correlation can reliably predict the performance of MJDs on \emph{fine-tuning comparison}, as supported by visual investigations. 
All our code is available at \href{https://github.com/mainlp/MJD-Estimator}{https://github.com/mainlp/MJD-Estimator} for reproduction.

%% file: Sections/related_work.tex
\section{Related Work}\label{sec:related-work}

\paragraph{Human label variation}

Human label variation represents a phenomenon in which inherent disagreement exists in annotation due to genuine disagreement, subjectivity or simply because two (or more) views are plausible \cite{DBLP:conf/emnlp/Plank22}.
\citet{DBLP:journals/aim/AroyoW15} propose that disagreement is not noise but signal, which is giving us information, as human labels are bound to be scarce yet at the same time critical as they provide human interpretations and values.
With substantial datasets, providing many judgments by high-quality coders for each item, training directly with soft labels achieved better results than training from aggregated or even gold labels \cite{DBLP:journals/jair/UmaFHPPP21}.
For example, \citet{fornaciari-etal-2021-beyond} find that the soft-label prediction auxiliary task reduces the penalty for errors on ambiguous entities and thereby mitigates overfitting.
Thus, \citet{DBLP:journals/tacl/PavlickK19} argue for a refined evaluation objective that requires models to explicitly capture the full distribution of plausible human judgments.

\paragraph{Human label variation in NLI}

For NLI---the task to determine whether a given premise entails, contradicts, or is neutral towards a target hypothesis---there exist several datasets addressing human label variation (HLV). 
On the one hand, 
Variation NLI \cite{DBLP:journals/tacl/PavlickK19} and Chaos NLI ~\cite{DBLP:conf/emnlp/NieZB20} collect annotations from 50 and 100 crowd workers for human judgment distributions (HJDs). 
On the other, Live NLI \cite{DBLP:conf/emnlp/JiangTM23} and VariErr NLI \cite{DBLP:journals/corr/abs-2403-01931} are annotated by only 5 or 4 linguists, but add textual explanations to their NLI labels. While these works have used NLI with explanations, their goals differ: either to unravel more on reasons for disagreement from the annotators \cite{DBLP:conf/emnlp/JiangTM23} or to use explanations to facilitate separation of plausible variation from annotation errors~\cite{DBLP:journals/corr/abs-2403-01931}.  However, little research has been done on estimating HJDs from a few labels and explanations.

\paragraph{Human explanations \& LLM estimators}

Human explanations, particularly ecologically valid ones where the same annotator provides both the label and explanation \cite{DBLP:conf/emnlp/JiangTM23}, 
are effective in improving LLMs' performance \cite{DBLP:conf/nips/Wei0SBIXCLZ22, DBLP:conf/emnlp/LampinenDCMTCMW22}.
\citet{DBLP:journals/corr/abs-2305-14770} investigate the levels of missing information in a provided answer for a target question in the context of a given article.
They use LLMs to rescale the coarse-grained (4-level) labels and accompanying explanations to a 100-point scale and compare them with manually annotated HJDs.  
\citet{pavlovic-poesio-2024-effectiveness} use GPT-3.5 to estimate HJDs directly, but in contrast to us focus on datasets with inherently subjective NLP tasks provided by the SemEval 2023 shared task 11~\cite{leonardelli-etal-2023-semeval}. Their exploration shows that
GPT-3.5 tends to produce distributions not well aligned with HJDs for subjective tasks. 

LLMs are also employed on the two NLI datasets with explanations. 
\citet{DBLP:conf/emnlp/JiangTM23} prompt GPT-3 to predict labels and generate explanations on LiveNLI items. 
They show that the predict-then-explain (post-prediction explanation) strategy significantly outperforms explain-then-predict (chain-of-thought prompting).
They also observe through qualitative analysis that over half of the LLM-generated explanations lack informativeness, i.e., only restate the premise/hypothesis.
\citet{DBLP:journals/corr/abs-2403-01931} ask GPT-3.5 and GPT-4 to judge the probability of whether individual explanations in VariErr make sense for the corresponding NLI labels.
Results show that GPT-4 outperforms traditional error detection methods, but the latter only evaluate the labels without factoring in the explanations, leaving several questions open.

\citet{DBLP:conf/emnlp/LeeAT23} is the closest to our work.
They propose an LLM Distribution Estimator that reads premise-hypothesis pairs and generates label distributions.
However, their MJDs align poorly with HJDs.
Moreover, explanation-contained NLI datasets were not yet available when \citet{DBLP:conf/emnlp/LeeAT23} was published.
Our paper bridges the recently released detailed explanation annotations with LLMs to estimate HJD on NLI efficiently.

%% file: Sections/method.tex
\section{LLMs to Estimate HJDs}
\label{sec:estimate-hjd-explanation-llm}

\newcolumntype{P}[1]{>{\ttfamily\raggedright\arraybackslash}p{#1}}

As LLM outputs are typically in text form and fluctuate based on inputs and parameters, obtaining model judgment distributions (MJDs) directly from the outputs to approximate human judgment distributions (HJDs) is challenging. 
We propose LLM prompts with multiple-choice questions (\S\ref{subsec:estimate-prompts}),\footnote{In preliminary experiments, we found that directly asking LLMs for MJDs was challenging. We also found another interesting prompt format in Appendix~\ref{app:assistant}.} 
to illustrate how we estimate MJD using first-token probability (\S\ref{subsec:estimate-first-token}) and reduce biases via permutations of choice options and explanations (\S\ref{subsec:estimate-bias}).

\subsection{Prompt Types}\label{subsec:estimate-prompts}

To facilitate asking LLMs to estimate MJDs, we transform the NLI problem into a multiple-choice question answering (MCQA) prompt, selecting one answer from the three options \texttt{[A,B,C]}. The prompts are shown in detail in Table \ref{tab:prompts} in Appendix \ref{app:prompt}. 
We design three prompt types: a base prompt ``without explanations'', a prompt ``with explanations'', and one ``with explicit explanations'' which contains both label and explanation. 

\paragraph{Without explanation} Our base prompt informs LLMs about the NLI task, provides the premise and hypothesis of NLI instances and instructs the LLMs to choose an NLI label via a MCQA format.
We constrain the label selection space by asking LLMs to ``select ONE of the listed options'' and restrict the initial letter of the output to one of \texttt{[A,B,C]} by instructing ``start your answer with a single letter.''

\paragraph{With explanations}
This prompt incorporates human explanations of label choices as ``comments'' without disclosing the annotators' chosen labels directly.
Specifically, we place these comments in the prompt after the hypothesis and premise but before the MCQA part. 

\paragraph{With explicit explanations} This prompt type reveals the NLI labels in addition to the corresponding explanations to the LLMs.
Our preliminary experiments found that LLMs sometimes misinterpret phrases that support a Neutral label (e.g., ``do not mention'' or ``cannot explain'') as supporting a Contradiction label. 
Therefore, we include a prompt type where the corresponding explanation by the annotator 
is appended with the phrase ``so I choose X'' to clarify the intended NLI label.

\subsection{First-token Probability}\label{subsec:estimate-first-token}

Conditioned by the prompts above, we next map LLMs' output
from \texttt{[A,B,C]} to probabilities as MJDs.
In particular, we set up a one-to-one mapping
$f \colon O\rightarrow L$ from the option set $O$ to the label space $L$, where $O=\{A,B,C\}$ and $L=\{Entailment,Neutral,Contradiction\}$ with permutations on both $O$ and $L$ (cf. \S\ref{subsec:estimate-bias}).

Denote the text output of LLMs as a list of words
$\bm{w} = [w_1,w_2,...,w_k], w_i \in V$ where $k$ is the length of the text output and $V$ is the vocabulary used for LLMs. We extract the logits of the first-token $w_1$ before the decoding process as
$\bm{s}_{w_1} = [s_{w_1},s_{w_2},...,s_{w_k},s_{w_{k+1}},...,s_{w_n}], w_i \in V$ where $n$ is the vocabulary size.
As shown in Figure~\ref{fig:overall}, we only use part of the first-token logits
$\bm{s}^{O}_{w_1} = [s_{A},s_{B},s_{C}]$ which present the distribution scores of the option set $O$.
To transform $\bm{s}^{O}_{w_1}$ into a probability distributions $\bm{p}^{O}$, we utilize normalization and softmax (with temperature \(\tau\)) functions:\footnote{The default $\tau$ is set to be 20; more studies in Appendix \ref{app:discuss_t}. 
}

\begin{equation}
p^{O}_{\text{norm}} (j) = \frac{{s}_j}{\sum_{j}^{|O|} {s}_j} ,
\quad
\end{equation}
\begin{equation}
p^{O}_{\text{sfmax}} (j) = \frac{\exp( s_j / \tau )}{\sum_{j}^{|O|} \exp({s}_j/ \tau)} ,
\quad
\end{equation} \label{equation2}

Finally, we obtain the model judgement distribution $\bm{p}^{L}$ 
through the mapping function $f$:

\begin{equation}
\bm{p}^{L} = f(\bm{p}^{O}) .
\quad
\end{equation}

\begin{table*}[t]
\centering
\resizebox{\textwidth}{!}{
\begin{tabular}{lccccl}
\toprule
Datasets & |Samples| & |Valid overlap| & |Annotators| & Explanations & Description \\
\midrule
MNLI \cite{DBLP:conf/naacl/WilliamsNB18} & 433k & 341 & 1 or 5 & \xmark & Majority with single label, subset with 5.
\\
VariErr NLI \cite{DBLP:journals/corr/abs-2403-01931} & 500 & 341 & 4 & \cmark & Ecologically valid explanations.
\\
MNLI subset of Chaos NLI \cite{DBLP:conf/emnlp/NieZB20} & 1,599 & 341 &100 & \xmark & Human judgment distributions.
\\
\bottomrule
\end{tabular}}
\caption{Datasets statistics. Numbers represent either the sample count or the annotator count.}\label{tab:datasets}
\end{table*}

\subsection{Bias Consideration}\label{subsec:estimate-bias}

Previous studies \citep[e.g.,][]{DBLP:journals/corr/abs-2306-07951,DBLP:journals/corr/abs-2309-03882,DBLP:journals/corr/abs-2311-04076} reveal that LLMs are biased when processing multiple-choice questions, such as preferring the first option A.
To address this, we shuffle the option set $O$ of the mapping relationship $f$,
totaling $A\binom {3}{3}=6$ permutations.
We ask LLMs to process all permutations, thus averaging all MJDs obtained to mitigate bias.
Namely, 
each of the three NLI labels has been mapped to option A,
reducing biases caused by the initial letters of the options.
Moreover, the order of multiple explanations may also exert unequal influence on LLMs, as shown in a case study in Appendix~\ref{app:case_study}. 
To address this bias, we adopt two ways to feed explanations (i.e., ``comments'') to the LLMs: ``serial'' and ``parallel'' modes.

\paragraph{Serial mode}
We input all explanations together to LLMs as in the prompt shown in Table~\ref{tab:prompts}, asking LLMs to process them at once. 
To mitigate the bias caused by the position of the explanations in the prompt, we shuffle the order of all $m$ explanations with full permutations $A\binom {m}{m}$ and use the average output as the model's final answer.

\paragraph{Parallel mode}
We feed one explanation (i.e., ``comment'') at a time under the ``parallel'' mode to the prompt.
Namely, to process $m$ explanation annotations on an NLI item, we create $m$ prompts and ask LLMs to handle them separately.
We then average the $m$ outputs to obtain the final MJD.
Although this approach prevents LLMs from obtaining an overall impression of all explanations, it allows them to focus more on interpreting each explanation and significantly reduces the computational cost from $A\binom {m}{m}$ (``serial'') to $m$ (``parallel'').

%% file: Sections/experiment_setup.tex
\section{Experimental Setup}\label{sec:experiment}

\subsection{Datasets}\label{subsec:experiment-datasets}

We experiment with two NLI datasets containing HLV, Chaos NLI \cite{DBLP:conf/emnlp/NieZB20} and VariErr NLI \cite{DBLP:journals/corr/abs-2403-01931}.
The former contains 1,599 MNLI \citep{DBLP:conf/naacl/WilliamsNB18} instances with HJDs collected from 100 crowd workers. In contrast, the latter contains 500 MNLI instances randomly sampled from Chaos NLI re-annotated with explanations and labels by 4 expert linguists.
Note that the source MNLI corpus is much larger than Chaos NLI and VariErr NLI,\footnote{MNLI contains 433k instances, but most of them with a single label, and only a small subset of the dev and test data were re-annotated by 4 additional annotators.} and is used only for pre-training of the fine-tuning experiments.
To ensure fairness and to facilitate follow-up analyses, 
we focus our comparisons on a subset of 341 VariErr items that receive exactly 4 explanation annotations.\footnote{In VariErr, each annotator can give one or more label-explanations pairs to an NLI item, and ``I don't know'' explanations are dropped after validation. 
Among 500 VariErr NLI items, 4 items received 1 validated explanation, 30 items 2, 62 items 3, 341 items 4, 59 items 5, and 4 items 6.} 
Table~\ref{tab:datasets} presents detailed data statistics. 

\subsection{Models}\label{subsec:experiment-models}
We utilized two open-source instruction-tuned LLMs:
Mixtral-8x7b-Instruct-v0.1 \cite{DBLP:journals/corr/abs-2401-04088} and Llama3-Chat-70b \cite{llama3}.
We adopt the original chat templates for both models and set the parameter \textit{do\_sample=False} in decoding to ensure consistent outputs for the same input.
Discussion regarding the risk of data leakage is elaborated in Appendix~\ref{app:leakage}.

\subsection{Distribution Comparison}\label{subsec:experiment-distribution-comparison}
Our first experiment compares the LLM-derived MJDs to the HJDs. To derive MJDs, we fed LLMs with the three prompt types exemplified in Table~\ref{tab:prompts} in the Appendix using the  341 VariErr instances. We then compare the resulting MJD to the HJDs of the corresponding Chaos NLI instances. 
We investigate these distribution differences between
humans and LLMs at the instance level following prior work \citep{DBLP:conf/emnlp/NieZB20,chiang-lee-2023-large, DBLP:conf/emnlp/LeeAT23}:  Kullback-Leibler Divergence (KL, \citealt{kullback1951information}) and Jensen-Shannon Distance (JSD, \citealt{DBLP:journals/tit/EndresS03}). In addition, we follow~\citet{baan-etal-2022-stop} to measure the human Distribution Calibration Error, namely measured as Total Variation Distance (TVD, \citealt{DBLP:books/daglib/0018090}) between MJDs and HJDs. 
MNLI single labels were transformed into one-hot vectors to compute the metrics.

\subsection{Fine-tuning Comparison}\label{subsec:experiment-finetuning-distribution}
Our second experiment investigates how well the resulting MJDs approximate human labels for model training.  
To do so, we compare the generated MJDs and original HJDs to annotated labels of the parallel instances in MNLI, VariErr NLI and Chaos NLI, for fine-tuning smaller language models, namely,  BERT \cite{DBLP:conf/naacl/DevlinCLT19} and RoBERTa \cite{DBLP:journals/corr/abs-1907-11692} base.
These models were first fine-tuned on the large single-labelled MNLI dataset to learn the generic NLI task. We then few-shot-tune them on the HJDs or MJDs above; see Appendix~\ref{app:imple} for fine-tuning details.

To evaluate the resulting classifiers, we split the remaining 1,258 MNLI instances from Chaos NLI that do not overlap with VariErr NLI into the development and test sets. 
We use KL and weighted F1 scores as evaluation metrics between the outputs of the fine-tuned models and HJDs from Chaos NLI. All metrics are detailed in Appendix~\ref{app:metrics}.

%% file: Sections/results.tex
\begin{table}[t]
\centering
\resizebox{0.48\textwidth}{!}{
\begin{tabular}{lccc}
\toprule
\textbf{Distributions}\textbackslash{}\textbf{Metrics}                                           & \textbf{KL $\downarrow$} & \textbf{JSD $\downarrow$} & \textbf{TVD $\downarrow$}       \\
\midrule
\multicolumn{4}{l}{\textit{Baseline}}   \\
\midrule
Chaos NLI                                                                               & 0    &0 &0  \\
\midrule
MNLI single label                                                                               & 9.288          & 0.422  &0.435                       \\
MNLI distributions                                                                               & 1.242          & 0.281   &0.295                \\
VariErr distributions                                                                           & 3.604           & 0.282   &0.296              \\
\midrule
Uniform distribution  & 0.364 & 0.307 & 0.350 \\
\midrule
\multicolumn{4}{l}{\textit{MJDs from Mixtral}}   \\
\midrule
$\bm{p}_{\text{norm}}$ of Mixtral                                                                  & 0.433        & 0.291   & 0.340         \\
+ ``serial'' explanations                                                                              & 0.407          & 0.265     &0.306          \\
+ ``serial'' explicit explanations                                                                                     & 0.382          & {0.246} &{0.286}           \\
+ ``parallel'' explanations                                                                              & {0.339 }         & {0.258}     &{0.295}         \\
+ ``parallel'' explicit explanations                                                                    & \textbf{0.245}          & \textbf{0.211}    &\textbf{0.239}     \\
\midrule
$\bm{p}_{\text{sfmax}}$ of Mixtral                                                                             & 0.434          & 0.292     &0.342                   \\
+ ``serial'' explanations                                                                                       & 0.349          & 0.258       &0.296         \\
+ ``serial'' explicit explanations                                                                                                  & {0.305}          & {0.235}     &{0.269}                       \\
+ ``parallel'' explanations                                                                                        & {0.310}          & {0.255}     & {0.290}          \\
+ ``parallel'' explicit explanations                                                                                 & \textbf{0.217}          & \textbf{0.208}    &\textbf{0.232}            \\
\midrule
\multicolumn{4}{l}{\textit{MJDs from Llama3}}   \\
\midrule
$\bm{p}_{\text{norm}}$ of Llama3                                                                  & 0.259          & 0.262       &0.284          \\
+ ``serial'' explanations                                                                              & 0.255          & 0.259   &0.281          \\
+ ``serial'' explicit explanations                                                                       & \textbf{0.235}          & \textbf{0.247}   &\textbf{0.266}                               \\
+ ``parallel'' explanations                                                                              & 0.257          & 0.261       &0.283        \\
+ ``parallel'' explicit explanations                                                                    & 0.243          & 0.253       &0.273        \\
\midrule
$\bm{p}_{\text{sfmax}}$ of Llama3                                                                           & 0.231          & 0.245    &0.260           \\
+ ``serial'' explanations                                                                                         & 0.226          & 0.243     &0.258               \\
+ ``serial'' explicit explanations                                                                                 & \textbf{0.212}          & \textbf{0.232}   &\textbf{0.245}                     \\
+ ``parallel'' explanations                                                                                      & 0.226          & 0.245      &0.260          \\
+ ``parallel'' explicit explanations                                                                              & 0.214          & 0.237     &0.254             \\
\bottomrule
\end{tabular}}
\caption{Distribution comparison results. MJDs and HJDs are compared on 341 overlapping 
instances.
}\label{tab:main-di}
\end{table}

\begin{table*}[t]
\centering
\resizebox{\textwidth}{!}{
\begin{tabular}{lccc|ccc}
\toprule
\multicolumn{1}{c}{\multirow{2}{*}{\textbf{Distributions}}}  & \multicolumn{3}{c|}{\textbf{BERT FT (dev / test)}} & \multicolumn{3}{c}{\textbf{RoBERTa FT (dev / test)}}\\ \cmidrule(lr){2-7} 
\multicolumn{1}{c}{}                         & \textbf{Weighted F1 $\uparrow$}                      & \textbf{KL $\downarrow$} &  \textbf{CE Loss $\downarrow$}   & \textbf{Weighted F1 $\uparrow$}  & \textbf{KL $\downarrow$}  &  \textbf{CE Loss $\downarrow$}     \\
\midrule
\multicolumn{7}{l}{\textit{Baseline}}   \\
\midrule
Chaos NLI train set                  & \textbf{0.626} / \textbf{0.646} & \textbf{0.074} / \textbf{0.077} & \textbf{0.972} / \textbf{0.974} & \textbf{0.699} / \textbf{0.650} & \textbf{0.061} / \textbf{0.067} & \textbf{0.932} / \textbf{0.943} \\
MNLI single label                      & 0.561          / 0.589          & 0.665          / 0.704          & 2.743          / 2.855          & 0.635          / 0.603          & 0.844          / 0.867          & 3.281          / 3.344          \\
MNLI distributions                     & 0.546          / 0.543          & 0.099          / 0.102          & 1.046          / 1.048          & 0.613          / 0.604          & 0.100          / 0.096          & 1.047          / 1.029          \\
VariErr distributions                  & 0.557          / 0.559          & 0.179          / 0.186          & 1.286          / 1.299          & 0.617          / 0.589          & 0.174          / 0.197          & 1.269          / 1.333          \\
\midrule
\multicolumn{7}{l}{\textit{MJDs from Mixtral}}   \\
\midrule
$\bm{p}_{\text{norm}}$ of Mixtral                                & 0.416          / 0.422          & 0.134          / 0.133          & 1.152          / 1.142          & 0.486          / 0.466          & 0.123          / 0.127          & 1.118          / 1.123          \\
+ ``serial'' explanations              & 0.443          / 0.454          & 0.145          / 0.141          & 1.183          / 1.166          & 0.509          / 0.514          & 0.128          / 0.128          & 1.132          / 1.126          \\
+ ``serial'' explicit   explanations   & 0.506          / 0.511          & 0.130          /0.130          & 1.139          / 1.132          & \textbf{0.569} / \textbf{0.572} & 0.114          / 0.122          & 1.091          / 1.107          \\
+ ``parallel'' explanations            & 0.404          / 0.428          & 0.134          / 0.131          & 1.150          / 1.136          & 0.483          / 0.502          & 0.123          / 0.122          & 1.118          / 1.109          \\
+ ``parallel'' explicit   explanations & \textbf{0.507} / \textbf{0.514} & \textbf{0.108} / \textbf{0.108} & \textbf{1.074} / \textbf{1.065} & 0.558          / 0.565          & \textbf{0.092} / \textbf{0.098} & \textbf{1.025} / \textbf{1.037} \\
\midrule
$\bm{p}_{\text{sfmax}}$ of Mixtral                                & 0.427          / 0.432          & 0.131          / 0.129          & 1.140          / 1.130          & 0.497          / 0.472          & 0.121          / 0.125          & 1.112          / 1.118          \\
+ ``serial'' explanations              & 0.452          / 0.462          & 0.121          / 0.118          & 1.113          / 1.096          & 0.506          / 0.525          & 0.110          / 0.109          & 1.078          / 1.069          \\
+ ``serial'' explicit   explanations   & 0.509          / \textbf{0.520} & 0.105          / 0.105          & 1.064          / 1.057          & \textbf{0.568} / 0.573          & 0.093          / 0.098          & 1.026          / 1.036          \\
+ ``parallel'' explanations            & 0.397          / 0.429          & 0.121          / 0.119          & 1.112          / 1.098          & 0.497          / 0.505          & 0.110          / 0.111          & 1.079          / 1.074          \\
+ ``parallel'' explicit   explanations & \textbf{0.522} / 0.517          & \textbf{0.095} / \textbf{0.095} & \textbf{1.035} / \textbf{1.026} & 0.567          / \textbf{0.576} & \textbf{0.082} / \textbf{0.087} & \textbf{0.994} / \textbf{1.003} \\
\midrule
\multicolumn{7}{l}{\textit{MJDs from Llama3}}   \\
\midrule
$\bm{p}_{\text{norm}}$ of Llama3                                  & 0.514          / 0.526          & 0.097          / 0.098          & 1.038          / 1.036          & 0.541          / 0.528          & 0.091          / 0.094          & 1.023          / 1.025          \\
+ ``serial'' explanations              & 0.574          / 0.574          & 0.096          / 0.097          & 1.037          / 1.033          & 0.618          / 0.601          & 0.091          / 0.093          & 1.020          / 1.022          \\
+ ``serial'' explicit   explanations   & 0.578          / 0.574          & \textbf{0.091} / \textbf{0.092} & \textbf{1.022} / \textbf{1.018} & 0.634          / 0.598          & \textbf{0.085} / \textbf{0.088} & \textbf{1.003} / \textbf{1.006} \\
+ ``parallel'' explanations            & 0.573          / 0.582          & 0.098          / 0.098          & 1.041          / 1.038          & 0.636          / 0.598          & 0.093          / 0.095          & 1.026          / 1.028          \\
+ ``parallel'' explicit   explanations & \textbf{0.582} / \textbf{0.586} & 0.094          / 0.095          & 1.030          / 1.026          & \textbf{0.639} / \textbf{0.620} & 0.089          / 0.091          & 1.014          / 1.016          \\
\midrule
$\bm{p}_{\text{sfmax}}$ of Llama3                                  & 0.528          / 0.524          & 0.091          / 0.093          & 1.023          / 1.021          & 0.546          / 0.535          & 0.085          / 0.089          & 1.005          / 1.009          \\
+ ``serial'' explanations              & 0.567          / 0.576          & 0.091          / 0.091          & 1.021          / 1.016          & 0.626          / 0.608          & 0.082          / 0.086          & 0.996          / 1.000          \\
+ ``serial'' explicit   explanations   & \textbf{0.585} / 0.568          & \textbf{0.086} / \textbf{0.087} & \textbf{1.008} / \textbf{1.004} & \textbf{0.646} / 0.610          & \textbf{0.077} / \textbf{0.081} & \textbf{0.981} / \textbf{0.987} \\
+ ``parallel'' explanations            & 0.584          / \textbf{0.583} & 0.092          / 0.093          & 1.024          / 1.020          & 0.643          / 0.611          & 0.085          / 0.089          & 1.004          / 1.008          \\
+ ``parallel'' explicit   explanations & 0.581          / 0.578          & 0.088          / 0.089          & 1.014          / 1.010          & 0.645          / \textbf{0.621} & 0.081          / 0.085          & 0.993          / 0.996             
\\
\bottomrule
\end{tabular}}
\caption{Results of fine-tuning comparison on Chaos NLI dev/test set.
The KL and Cross-Entropy (CE) Loss reflected the distance between distributions, whereas Weighted F1 reflected the capability in handling NLI problems. }\label{tab:main-ft}
\end{table*}

\section{Results}\label{sec:results}

\subsection{Distribution Comparison}\label{subsec:results-distribution-comparisons}

Table~\ref{tab:main-di} presents the distribution comparison results.
Firstly, we analyze baseline HJDs and observe that
the MNLI single-label data (i.e., with no HLV) is the farthest from Chaos NLI's HJD, followed by MNLI and VariErr distributions. 
We add a distribution comparison to the uniform distribution as a sanity check to understand the obtained MJDs.

When comparing MJDs, we observe that for the ``without explanation'' prompt, Llama3 is closer than Mixtral to Chaos NLI HJD. 
However, both models benefit from adding explanations, i.e., their MJDs gradually get closer to the HJD. 
The two LLMs reach similar scores overall, exceeding those of MNLI and VariErr distributions regardless of whether the transformation method is $\bm{p}_{\text{norm}}$ or $\bm{p}_{\text{sfmax}}$.
Mixtral benefits more from the additional information (KL/JSD/TVD drop more from its base), and the best setup uses explicit explanations.

\begin{figure}[t]

        \centering
        \includegraphics[width=\columnwidth]{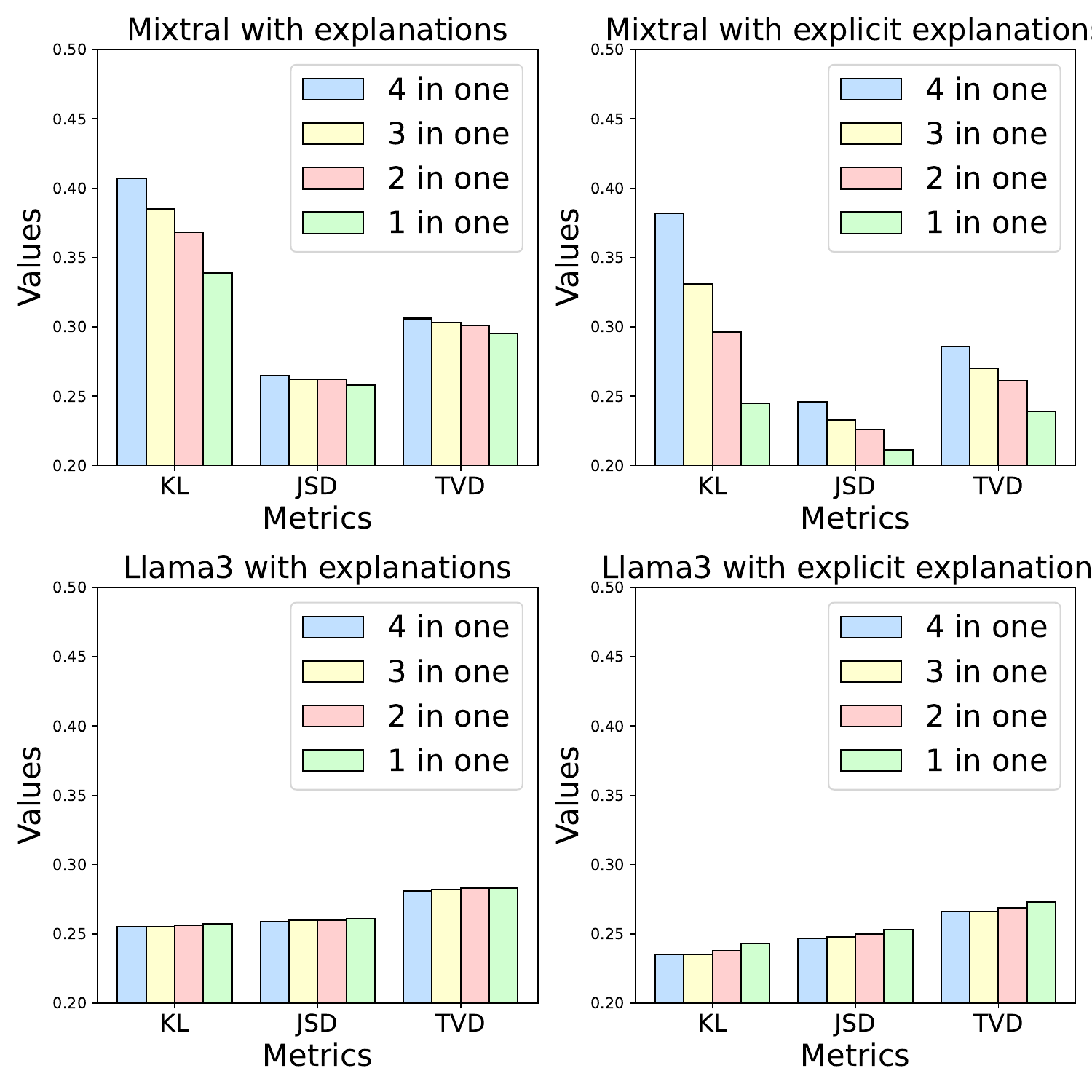}
        \caption{Distribution comparison results.
        ``n in one'' denotes the way LLMs process $n$ explanations at a time.}
        \label{fig:ablation}
    
\end{figure}

\paragraph{Parallel vs serial}
\label{sec:serial}
Regarding bias considerations, the two LLMs exhibit slightly different patterns. 
``Serial'' mode is better for Llama, while ``parallel'' suits Mixtral. 
``Serial'' mode intuitively feels better because it provides multiple explanations at the same time, allowing models to estimate each label relative to all other labels and explanations. For example, LLMs may discriminate which explanation is more convincing and thus favor the label indicated by that explanation.
We conduct an ablation study by adding 4, 3, 2, or 1 explanations at a time to LLM prompts.
However, Figure~\ref{fig:ablation} shows that adding more explanations to a prompt makes the Mixtral MJDs less similar to HJD but more for Llama3 MJDs.
The distribution similarity between Mixtral MJDs and HJDs seems to gradually decrease as the number of input explanations increases, whereas Llama3 shows almost no fluctuation and even performs better with longer texts.
We hypothesize that Llama3 is better at longer prompts.
More detailed scores are listed in Table~\ref{tab:ablation} in Appendix~\ref{app:ablation}.

\subsection{Fine-tuning Comparison}\label{subsec:results-finetuning-comparisons}

We present results on fine-tuning smaller models and comparing their MJDs to the HJDs on the held-out Chaos NLI dev and test sets.
As comparison to LLM-predicted HJDs, we also train models on the existing datasets; see
Table~\ref{tab:main-ft} for the results. All detailed scores of fine-tuning comparison are in Table~\ref{tab:main-ft-eval} and Table~\ref{tab:main-ft-test} in Appendix~\ref{app:detail}.

Models trained on the Chaos NLI train set perform best, noticeably higher than MNLI and VariErr models. 
Regarding the KL and CE Loss metrics, both LLMs demonstrate strong approximation performance, approaching Chaos NLI more closely than MNLI or VariErr NLI. Mixtral gets slightly better KL and CE Loss on ``parallel'', while Llama3 wins on ``serial'', mirroring results in \S\ref{subsec:results-distribution-comparisons}.

However, fine-tuned LLM models show divergent results on F1.
Overall, adding explicit explanations contributes to the best models.
Llama3 improves fine-tuning results by yielding closer MJD to HJD and achieving better F1 scores than MNLI/VariErr HJDs.
In contrast, while Mixtral only achieves slightly worse results in distribution comparison, it is much inferior in F1 score, even below that of MNLI/VariErr HJDs.
The next section investigates this discrepancy in LLM performances between distribution and fine-tuning comparisons. 

\section{Analysis and Discussion}\label{sec:analysis}

We observe from \S\ref{sec:results} that even though Llama3 and Mixtral are equally similar to Chaos NLI HJDs in distribution comparisons (see Table~\ref{tab:main-di}), their fine-tuned F1 scores differ, with Llama3 MJDs achieving a higher F1 than MNLI/VariErr HJDs while Mixtral being lower (see Table~\ref{tab:main-ft}).  We further inspect these distributions visually (\S\ref{subsec:analysis-visualization}) and using distance correlation (\S\ref{subsec:analysis-correlations}). We then explore the potential causes for the observed differences (\S\ref{subsec:analysis-why-llama-better}). Finally, we highlight the contributions of this paper and suggest future directions for reference (\S\ref{subsec:analysis-future}).

\subsection{Visualization}\label{subsec:analysis-visualization}

\begin{figure}[t]

        \centering
        \includegraphics[width=\linewidth]{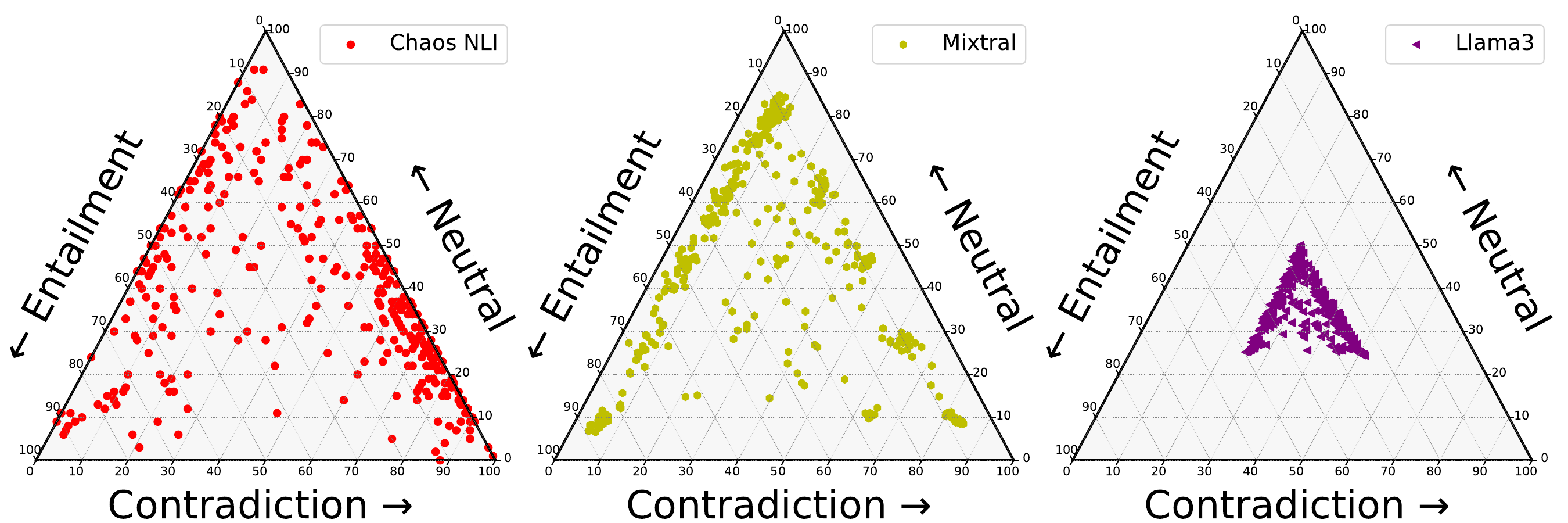}
        \caption{Visualization of distributions in ternary plot. Each point represents one of the 341 samples. }
        \label{fig:visual-distributions-2}
\end{figure}

\begin{figure}[t]

        \centering
        \includegraphics[width=\linewidth]{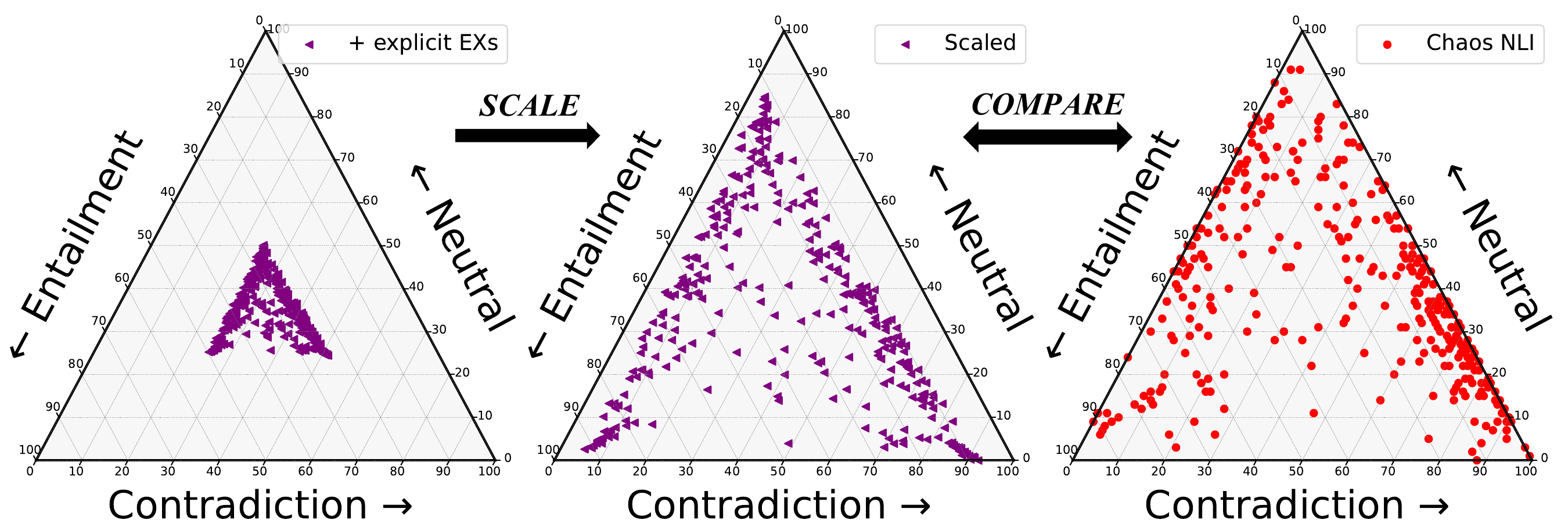}
        \caption{Zooming in (\textit{scale=3.3}) on Llama3 MJD.
        }
        \label{fig:visual-scale}
    
\end{figure}

\begin{figure}[t]

        \centering
        \includegraphics[width=\linewidth]{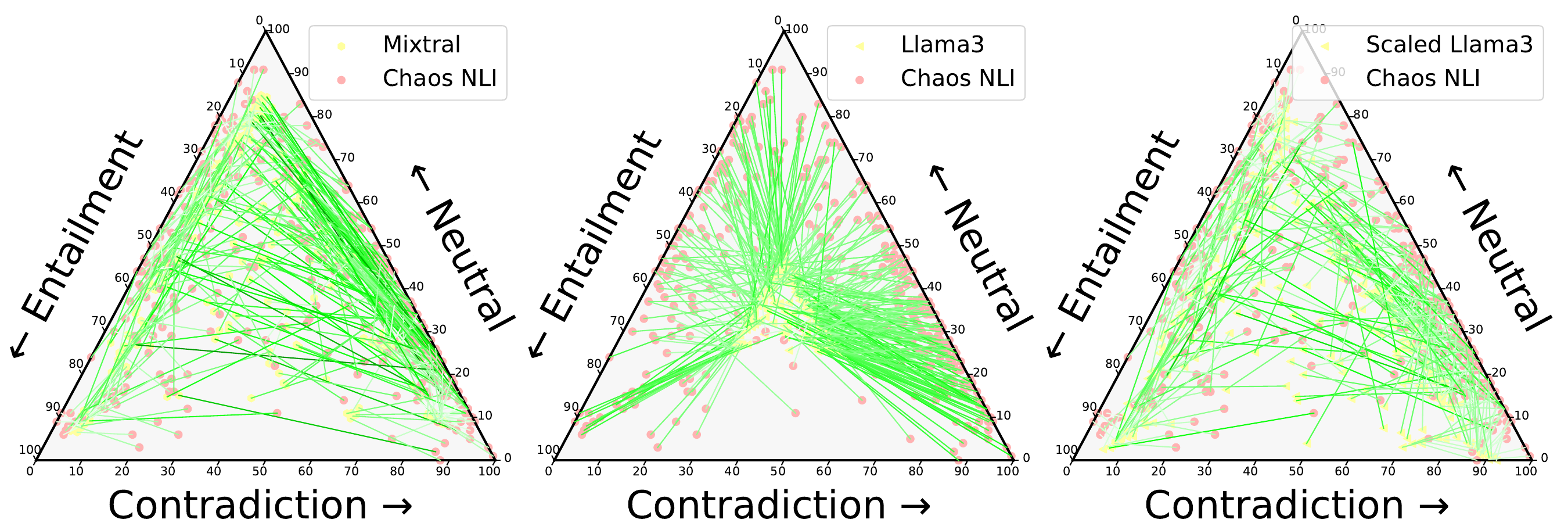}
        \caption{Error visualization. Corresponding sample points from the MJD and Chaos NLI HJD connected by a green line. Darker green means more distant. }
        \label{fig:visual-errors-withline}
    
\end{figure}

To inspect MJDs against HJD, we use the visualization tool by~\citet{gruber-etal-2024-labels} to plot each instance's distribution in the ternary plot. For consistency, we focus on a single setting with ``explicit explanations'' prompt, ``parallel'' mode, and $\bm{p}_{\text{norm}}$ transformation.
Figure~\ref{fig:visual-distributions-2} compares the Chaos NLI HJD to Mixtral and Llama3's MJDs. 
Interestingly, Llama3 and Mixtral exhibit rather different clusters: Llama3 has an overall higher entropy with instances closer to the center, whereas Mixtral is seemingly closer to the Chaos NLI HJD. 
We also see that the original Chaos NLI HJD is slightly skewed towards Contradiction, i.e., the right side of the triangle, while Mixtral MJD is slightly skewed towards Entailment, i.e., the opposite left side.  
Figure~\ref{fig:visual-scale} further zooms in on Llama3 MJD and shows that Llama3 is slightly skewed towards the right side (Contradiction), more in line with Chaos NLI.

\paragraph{Pairwise distance plots} While triangle plots provide insights on the overall distributions, they lack information regarding \emph{how far} a particular instance (sample point) is in MJD, e.g., \textit{[E=0.4, N=0.5, C=0.1]} from its correspondence in the HJD. Therefore, we calculate and visualize pair-wise distances (or errors) 
between the corresponding sample points in MJDs and Chaos NLI HJDs.
Namely, if many samples ``moved'' dramatically from an HJD to an MJD, this is a worse MJD estimation. 

Figure~\ref{fig:visual-errors-withline} illustrates that Mixtral MJD has more substantial movements, i.e., erroneous estimates, than Llama3 MJD to Chaos NLI HJD. 
Figure~\ref{fig:visual} in Appendix \ref{app:visual} further provides pairwise error comparisons and shows that the absolute errors of Mixtral are scattered while Llama3 errors are more concentrated. 
These observations motivate us to look for a more suitable error measure in \S\ref{subsec:analysis-correlations}.

\subsection{Quantifying the Visual Observations:  Distance Correlation}\label{subsec:analysis-correlations}

Visualizations clearly show that, compared to Mixtral, the MJDs produced by Llama3 exhibit shapes more similar to HJDs, which corroborates Llama's superior performance in fine-tuning comparisons.
Given that the previous metrics used in \emph{distribution comparison} are focusing on instance-level, while visualizations represent the distribution of all data points, we propose to further evaluate MJDs against HJDs using a global-level measure, distance correlation (D.Corr, \citealt{szekely2007measuring}), to capture the differences between general distributions. 

We consider all the samples' soft labels from a dataset as a 3-D array. 
The D.Corr between the source dataset $X$ and the target dataset $Y$ is calculated as: 
\begin{equation}
    \operatorname{dCor}^2(X, Y)=\frac{\mathrm{dCov}^2(X, Y)}{\sqrt{\operatorname{dVar}^2(X) \operatorname{dVar}^2(Y)}} .
\end{equation}
where $\mathrm{dCov}^2(X, Y)$ means the distance covariance of the two arrays, and  $\operatorname{dVar}$ means the distance standard deviation of the array. 

This measure accounts for all soft labels across the dataset and thus should be considered a global measure of the overall datasets.
Results from Table~\ref{tab:new-metric} show that  Llama3 MJDs have a substantially higher distance correlation with Chaos NLI's HJD than Mixtral.
This further proves Llama3 is globally better aligned with the HJD than Mixtral and supports its better fine-tuning performances. 
Appendix~\ref{app:new_metric} shows more results with different temperatures $\tau$ and numbers of explanations.

\begin{table}[t]
\centering
\resizebox{0.37\textwidth}{!}{
\begin{tabular}{lcc}
\toprule
\textbf{Distributions}\textbackslash{}\textbf{Metrics}                                            & \textbf{D.Corr $\uparrow$}       \\
\midrule
Uniform distribution                                                                                 &0                         \\
MNLI single label &0.612                         \\
MNLI distributions                                                                                   &0.795         \\
VariErr distributions                                                                               &0.688      \\
\midrule
\multicolumn{2}{l}{\textit{MJDs from Mixtral}}   \\
\midrule
$\bm{p}_{\text{norm}}$ of Mixtral                                                           &0.609 \\
+ ``parallel'' explicit explanations                                                                &\textbf{0.719}  \\
\midrule
$\bm{p}_{\text{sfmax}}$ of Mixtral                                                                      &0.593     \\
+ ``parallel'' explicit explanations                                                                         &\textbf{0.709} \\
\midrule
\multicolumn{2}{l}{\textit{MJDs from Llama3}}   \\
\midrule
$\bm{p}_{\text{norm}}$ of Llama3                                                                &0.689  \\
+ ``parallel'' explicit explanations                                                           &\textbf{0.809}    \\
\midrule
$\bm{p}_{\text{sfmax}}$ of Llama3                                                                       &0.677      \\
+ ``parallel'' explicit explanations                                                           &\textbf{0.802}    \\
\bottomrule
\end{tabular}}
\caption{Distance Correlation (D.Corr) on distribution comparisons between MJDs and Chaos NLI HJD. 
A higher correlation indicates better performance.
}\label{tab:new-metric}
\end{table}

\subsection{Why did the Llama3 MJD Work Better than Mixtral's in Fine-tuning?}\label{subsec:analysis-why-llama-better}

To sum up, both the visualization in \S\ref{subsec:analysis-visualization} and distance correlations in \S\ref{subsec:analysis-correlations} provide additional insights into the differences between Mixtral's and Llama3's MJDs. 
We hypothesize that one advantage of the Llama3-generated soft labels is its \textbf{smoothness}. 
As the benefit of label smoothness for model training has been validated extensively in the past ~\cite{DBLP:conf/nips/MullerKH19,DBLP:conf/icml/WeiXCF0L22}, when we observe that Llama3's MJD has a smoother distribution, we then assume that smoother labels could be beneficial based on intuition from previous papers.
The overall higher entropy of the Llama3 non-scaled MJD (used for fine-tuning) has a regularizing effect similar to label smoothing. In other words, Llama3-generated MJD looks similar to the temperature-scaled (i.e., squished, within a certain range of $\tau$) version of Chaos NLI. 
Appendix~\ref{app:discuss_t} 
further visualizes $\bm{p}_{\text{sfmax}}$-transformed Llama3 MJDs with $\tau$ in [5, 10, 20]. 
From the D.Corr results in Table~\ref{tab:correlation} in Appendix~\ref{app:new_metric}, we indeed found that higher $\tau$-values (smoother label distributions) can lead to better scores with $\tau$ from 5, 10 to 20. This somehow supports our hypothesis that smoothness might be beneficial.

It is important to note that instance-level distribution distance measures such as KL and JSD cannot reliably predict 
models' performance when fine-tuned on the generated MJDs.
A lower KL divergence cannot guarantee that it contains more information for fine-tuning.
In the most extreme case, the KL divergence between a uniform random distribution and Chaos NLI is 0.364, much lower than the one-hot distribution on MNLI (0.665), while providing no helpful information for training.
By further inspecting the visualization result, we hypothesize that the second advantage of the Llama3-generated soft labels is the \textbf{shape} of the sample distribution.
That means the cross-sample dependency of the Llama3-generated soft labels is more similar to Chaos NLI than that of Mixtral. One way to capture this cross-sample dependency is to calculate a \textit{distance matrix} recording the distances between every sample pair in the MJD.

To compare the global level similarity between the generated and the target dataset, we thus proposed to use visual inspections and measure the distance correlation, which measures the distance matrix in the covariance.
We empirically show that distance correlation can better reflect the effectiveness of the generated MJDs for fine-tuning.
This led us to conclude that metrics like KL, JSD, and TVD, which measure the distance between distributions at the instance level, are better complemented by additional investigations on the shape of the resulting annotations using visualization techniques and global measures.

\subsection{Potential benefits beyond NLI} \label{subsec:analysis-future}

In this study we validate the effectiveness of our approach on NLI. Given that NLI requires different linguistic capabilities \citep{wang2018glue} such as, e.g., reasoning about word order, or understanding active/passive voice, we strongly believe that our findings generalize to other NLP tasks requiring similar capabilities. We aim to extend our study to other tasks in future work.

Furthermore, for a large number of tasks, constructing HLV datasets requires extensive annotation by crowd workers to obtain HJDs. This paper explores a method that only needs a small number of explanations (reasons why annotators made certain annotations) to approximate labor-intensive HJDs using LLMs. If datasets for more tasks include reasons for annotators' choices, our approach can be applied to approximate HJDs using LLMs for tasks such as sentiment analysis, stance detection and hate speech, etc., where there also exists disagreement and need for HJDs, see for example \citep{DBLP:conf/eacl/SandriLTJ23}. This would allow researchers to better explore the impact of HLV.

On a broader scale, research into HLV is akin to aligning current machine models with human values and addressing discrepancies in human viewpoints, different interpretations, and aligning them with machine judgment.

%% file: Sections/appendix.tex
\section{Prompt}
\label{app:prompt}

All prompts used in this paper are listed in Table~\ref{tab:prompts}.
We adopted the original chat templates for both LLMs to input the prompt.

\begin{table}[t]
\scriptsize
\centering
\begin{tabular}{P{0.09\textwidth} | P{0.33\textwidth}}
\toprule 
\multicolumn{1}{l|}{\textbf{Type}} & \multicolumn{1}{l}{\textbf{General Instruction Prompt}} \\
\midrule
Without explanations & \textbf{"role": "user", "content"}:  \newline  Please determine whether the following Statement is true (entailment), undetermined (neutral), or false (contradiction) given the Context below and select ONE of the listed options and start your answer with a single letter. \newline Context: \{promise\} \newline Statement: \{hypothesis\} \newline A. Entailment \newline B. Neutral \newline C. Contradiction. \newline Answer:                                 \\
\midrule  
With explanations & \textbf{"role": "user", "content"}:  \newline  Please carefully and fairly base your selection on the comments below to determine whether the following Statement is true (entailment), undetermined (neutral), or false (contradiction) given the Context below and select ONE of the listed options and start your answer with a single letter. \newline Context: \{promise\} \newline Statement: \{hypothesis\} \newline Comment 1: \{explanation 1\} \newline Comment 2: \{explanation 2\}\newline ... \newline A. Entailment \newline B. Neutral \newline C. Contradiction. \newline Answer:                                 \\
\midrule
With explicit explanations & \textbf{"role": "user", "content"}:  \newline  Please carefully and fairly base your selection on the comments below to determine whether the following Statement is true (entailment), undetermined (neutral), or false (contradiction) given the Context below and select ONE of the listed options and start your answer with a single letter. \newline Context: \{promise\} \newline Statement: \{hypothesis\} \newline Comment 1: \{explanation 1\}, so I choose \{label 1\} \newline Comment 2: \{explanation 2\}, so I choose \{label 2\} \newline ... \newline A. Entailment \newline B. Neutral \newline C. Contradiction. \newline Answer:                                 \\
\bottomrule
\end{tabular}

\caption{Instruction prompt of different types to transform NLI into a multi-choice question format.}
\label{tab:prompts}
\end{table}

\section{Experimental Implementation}
\label{app:imple}

We first fine-tuned BERT-base-uncased~\cite{DBLP:conf/naacl/DevlinCLT19} and Roberta-base~\cite{DBLP:journals/corr/abs-1907-11692} with the standard NLI training process on MNLI single labels~\cite{DBLP:conf/naacl/WilliamsNB18}, and then
fine-tuned them on the label distributions of MNLI, VariErr NLI~\cite{DBLP:journals/corr/abs-2403-01931} and Chaos NLI~\cite{DBLP:conf/emnlp/NieZB20}, as well as the MJDs generated by the LLMs. 
We used cross-entropy as a loss function for soft-label training.
For validation on the dev set, we measured the distribution distance by calculating KL divergence and cross-entropy between the logits of the model and the soft label. 
We also measured the prediction performance by calculating the F1 score using the largest logits of the model against the majority-voted label.
We selected the model with the best macro-F1 score performance on the dev set for final testing.
Detailed hyperparameter choices are listed in Table \ref{tab:hyperpara}.
Fine-tuning was conducted with NVIDIA A100 80GB within several hours.

\begin{table}[ht]
    \centering
    \begin{adjustbox}{width={0.35\textwidth},keepaspectratio}%

    \begin{tabular}{l r}
        \toprule 
        \textbf{Hyperparameter} & \textbf{Our Model} \\
        \midrule
        Learning Rate Decay & Linear \\
        Weight Decay & 0.0 \\
        Optimizer & AdamW \\
        Adam $\epsilon$ & 1e-8 \\
        Adam $\beta_{1}$ & 0.9 \\
        Adam $\beta_{2}$ & 0.999 \\
        Warmup Ratio & 0\% \\
        Learning Rate & 2e-5 \\
        Batch size & 4 \\
        Num Epoch & 5\\
        \bottomrule
    \end{tabular}
    \end{adjustbox}

    \caption{Hyperparameter used for fine-tuning BERT and RoBERTa models. }
    \label{tab:hyperpara}
\end{table}

\section{Metrics}
\label{app:metrics}

\paragraph{Kullback-Leibler Divergence}
Kullback-Leibler divergence, often referred to as KL divergence (KL), is a measure of how one probability distribution diverges from a second, reference probability distribution \cite{kullback1951information}. 
It is useful for capturing the relative entropy or information loss when approximating one distribution with another.
It is a non-symmetric measure of the difference between two probability distributions \(P\) and \(Q\).

KL divergence is very sensitive to differences between the two distributions. If there are points where one distribution assigns a high probability and the other assigns a low probability, KL divergence will highlight these differences significantly.

For discrete probability distributions \(P\) and \(Q\):

\begin{equation}
D_{\text{KL}}(P | Q) = \sum_{x \in \mathcal{X}} P(x) \log \frac{P(x)}{Q(x)} ,
\end{equation}

For continuous probability distributions, the sum is replaced by an integral:

\begin{equation}
D_{\text{KL}}(P | Q) = \int_{-\infty}^{\infty} p(x) \log \frac{p(x)}{q(x)} \, dx ,
\end{equation}

In this paper, we adopted the discrete version, as our target distributions are 3-dimensional probabilities.
KL divergence is asymmetric (\( D_{\text{KL}}(P | Q) \neq D_{\text{KL}}(Q | P) \)), which can be beneficial when the distributions have a clear direction of reference or when one distribution is considered the true distribution and the other is an approximation. Thus, we set Chaos NLI HJD as the true distribution \(P\), and MJDs as the approximation \(Q\).

\paragraph{Jensen-Shannon Distance}

Jensen-Shannon distance (JSD) is a symmetric and smoothed version of the KL divergence~\cite{DBLP:journals/tit/EndresS03}. 
Unlike KL divergence, Jensen-Shannon distance is symmetric (\( D_{\text{JS}}(P | Q) = D_{\text{JS}}(Q | P) \)). This makes it suitable when there is no inherent reference direction between the two distributions.
Jensen-Shannon distance is always finite and bounded between 0 and 1, making it easier to interpret and compare.
By averaging the two distributions, it mitigates the impact of any extreme values, providing a more stable measure of similarity.

For discrete probability distributions \(P\) and \(Q\), let \(M = \frac{1}{2}(P + Q)\). The Jensen-Shannon divergence is defined as:

\begin{equation}
D_{\text{JS}}(P | Q) = \sqrt{\frac{(D_{\text{KL}}(P | M) +  D_{\text{KL}}(Q | M))}{2} } ,
\end{equation}

Same as KL, we also set HJD from Chaos NLI as $P$, while MJDs as $Q$.

\paragraph{Total Variation Distance}

Total Variation distance (TVD) is a measure of the maximum difference between the probabilities assigned to the same event by two different probability distributions~\cite{DBLP:books/daglib/0018090} . It provides a way to quantify the difference between two distributions.

TV distance can be interpreted as the maximum proportion of the distribution that needs to be altered to transform one distribution into the other. This makes it an intuitive measure of overall dissimilarity.
Unlike KL divergence and JS distance, which involve logarithms and averages, TV distance is based on absolute differences. This can be particularly useful when you need a straightforward measure of discrepancy.
TV distance is robust to small changes in probability values, making it a reliable measure when comparing distributions that may have minor variations.

For discrete probability distributions \(P\) and \(Q\):

\begin{equation}
D_{\text{TV}}(P, Q) = \frac{1}{2} \sum_{x \in \mathcal{X}} |P(x) - Q(x)| ,
\end{equation}

For continuous probability distributions, the sum is replaced by an integral:

\begin{equation}
D_{\text{TV}}(P, Q) = \frac{1}{2} \int_{-\infty}^{\infty} |p(x) - q(x)| \, dx ,
\end{equation}

In this paper, we still used the discrete version, as same as KL. We set HJD from Chaos NLI as $P$, while MJDs as $Q$.

\paragraph{Weighted F1 Score}

The weighted F1 score is an extension of the standard F1 score that accounts for class imbalance in multi-class classification problems. In multi-class classification, different classes can have varying frequencies, and the weighted F1 score adjusts for this by giving more importance to classes that have more instances.

The F1 score is the harmonic mean of precision and recall:

\begin{equation}
\text{F1 Score} = 2 \times \frac{\text{Precision} \times \text{Recall}}{\text{Precision} + \text{Recall}} ,
\end{equation}

\noindent where precision (P) is the proportion of true positives among all predicted positives:
    \begin{equation}
    \text{Precision} = \frac{TP}{TP + FP} ,
    \end{equation}
    
\noindent and recall (R) is the proportion of true positives among all actual positives:
    \begin{equation}
    \text{Recall} = \frac{TP}{TP + FN} ,
    \end{equation}

In a multi-class setting, we calculate the F1 score for each class, then take a weighted average based on the number of true instances of each class. This alters macro F1 Score to account for label imbalance; it can result in an F-score that is not between precision and recall.

The formula for the weighted F1 score is:

\begin{equation}
\text{Weighted F1} = \frac{1}{N} \sum_{i=1}^{k} w_i \times F1_i .
\end{equation}

\noindent where \(k\) is the total number of classes, \(F1_i\) is the F1 score for class \(i\), \(w_i\) is the weight for class \(i\), which is proportional to the number of true instances of class \(i\) ($ w_i = \frac{n_i}{N} $, where \(n_i\) is the number of true instances of class \(i\), and \(N\) is the total number of instances across all classes.).

The weighted F1 score is particularly useful when dealing with imbalanced datasets because it adjusts the contribution of each class’s F1 score based on how common the class is. This prevents the model from being overly influenced by the performance on the majority class, which can be misleading in an imbalanced dataset. In this paper, the weighted F1 score is implemented with \emph{sklearn}\footnote{\href{https://scikit-learn.org/}{https://scikit-learn.org/}}.

\begin{table*}[htbp]
\centering
\resizebox{\textwidth}{!}{
\begin{tabular}{lcccccc}
\toprule
\multicolumn{1}{c}{\multirow{2}{*}{\textbf{Distributions}}} & \multicolumn{3}{c}{$p_{\text{norm}}$} & \multicolumn{3}{c}{$p_{\text{sfmax}}$} \\ \cmidrule(lr){2-7} 
\multicolumn{1}{c}{}                                               & \textbf{\textbf{KL $\downarrow$}} & \textbf{\textbf{JSD $\downarrow$}} & \textbf{\textbf{TVD $\downarrow$}}  &  \textbf{\textbf{KL $\downarrow$}} & \textbf{\textbf{JSD $\downarrow$}} & \textbf{\textbf{TVD $\downarrow$}}      \\
\midrule
Mixtral original                                                                  & 0.433        & 0.291   & 0.340              & 0.434          & 0.292     &0.342                           \\
4 explanations   at a time                                                                            & 0.407          & 0.265     &0.306           & 0.349          & 0.258       &0.296            \\
 3  explanations  at a time                                                                             & 0.385          & 0.262     & 0.303            & 0.338          & 0.257      &0.295        \\
 2  explanations   at a time                                                                            & 0.368          & 0.262     & 0.301            & 0.329          & 0.257       &0.293     \\
 1  explanations  at a time                                                                             & \textbf{0.339 }         & \textbf{0.258}     &\textbf{0.295}              & \textbf{0.310}          & \textbf{0.255}     & \textbf{0.290}             \\
\midrule
4  explicit explanations  at a time                                                                   & 0.382          & 0.246     &0.286             & 0.305          & 0.235      &0.269         \\
3  explicit explanations at a time                                                                    & 0.331          & 0.233      &0.270             & 0.275          & 0.226   &0.257      \\
 2  explicit explanations at a time                                                                    & 0.296          & 0.226      &0.261             & 0.254          & 0.221        &0.251             \\
1  explicit explanations   at a time                                                                  & \textbf{0.245}          & \textbf{0.211}    &\textbf{0.239}               & \textbf{0.217}          & \textbf{0.208}    &\textbf{0.232}            \\
\midrule
\midrule
Llama3 original                                                                  & 0.259          & 0.262       &0.284            & 0.231          & 0.245    &0.260              \\
 4 explanations at a time                                                                              & \textbf{0.255}          & \textbf{0.259}   &\textbf{0.281}              & 0.226          & \textbf{0.243}     &0.258                     \\
 3 explanations at a time                                                                              & 0.255          & 0.260     &0.282              & 0.225          & 0.243      &0.258                \\
2  explanations at a time                                                                              & 0.256          & 0.260       &0.283         & \textbf{0.224}          & 0.243         &\textbf{0.258}   \\
 1 explanation at a time                                                                                 & 0.257          & 0.261       &0.283            & 0.226          & 0.245      &0.260          \\
\midrule
 4 explicit explanations  at a time                                                                     & \textbf{0.235}          & \textbf{0.247}   &\textbf{0.266}                & 0.212          & 0.232      &0.245         \\
 3 explicit explanations  at a time                                                                   & 0.235          & 0.248      &0.266             & 0.209          & \textbf{0.232}        &\textbf{0.244}            \\
2 explicit explanations   at a time                                                                  & 0.238          & 0.250      &0.269           & \textbf{0.209}          & 0.233        &0.245          \\
1 explicit explanation at a time                                                                    & 0.243          & 0.253       &0.273            & 0.214          & 0.237     &0.254             \\
\bottomrule
\end{tabular}}
\caption{All results of ``serial''/``parallel''. Scores are compared with Chaos NLI HJD. Note that since all the results are averaged scores of $A\binom {m}{m}$ combinations (described in \S\ref{subsec:estimate-bias}), which means LLMs actually obtain all 4 explanations' information in every setting. ``Serial'' represents ``4 explanations at a time'', while ``parallel'' represents ``1 explanations at a time''. For the settings of 2 and 3, they are the transition form from ``serial'' to ``parallel''.}\label{tab:ablation}
\end{table*}

\begin{table*}[htbp]
\centering
\resizebox{\textwidth}{!}{
\begin{tabular}{lccccc|ccccc}
\toprule
\multicolumn{1}{c}{\multirow{2}{*}{\textbf{Distributions}}}  & \multicolumn{5}{c|}{\textbf{BERT FT}} & \multicolumn{5}{c}{\textbf{RoBERTa FT}}\\ \cmidrule(lr){2-11} 
\multicolumn{1}{c}{}                                               & \textbf{ACC. $\uparrow$} & \textbf{KL $\downarrow$} & \textbf{Weighted F1 $\uparrow$} & \textbf{Macro F1 $\uparrow$}  &  \textbf{CE Loss $\downarrow$}     & \textbf{ACC. $\uparrow$} & \textbf{KL $\downarrow$} & \textbf{Weighted F1 $\uparrow$} & \textbf{Macro F1 $\uparrow$}  &  \textbf{CE Loss $\downarrow$}     \\
\midrule
\multicolumn{10}{l}{\textit{Baseline}}   \\
\midrule
Chaos NLI train set                  & \textbf{0.628}          & \textbf{0.074}          & \textbf{0.626}          & \textbf{0.566}          & \textbf{0.972}          & \textbf{0.698}          & \textbf{0.061}          & \textbf{0.699}          & \textbf{0.659}          & \textbf{0.932}          \\
MNLI single label                      & 0.552          & 0.665          & 0.561          & 0.523          & 2.743          & 0.628          & 0.844          & 0.635          & 0.616          & 3.281          \\
MNLI distributions                     & 0.542          & 0.099          & 0.546          & 0.518          & 1.046          & 0.607          & 0.100          & 0.613          & 0.598          & 1.047          \\
VariErr distributions                  & 0.558          & 0.179          & 0.557          & 0.514          & 1.286          & 0.617          & 0.174          & 0.617          & 0.594          & 1.269          \\
\midrule
\multicolumn{4}{l}{\textit{MJDs from Mixtral}}   \\
\midrule             
$\bm{p}_{\text{norm}}$ of Mixtral             & 0.404          & 0.134          & 0.416          & 0.400          & 1.152          & 0.464          & 0.123          & 0.486          & 0.451          & 1.118          \\
+ ``serial'' explanations              & 0.459          & 0.145          & 0.443          & 0.452          & 1.183          & 0.506          & 0.128          & 0.509          & 0.502          & 1.132          \\
+ ``serial'' explicit   explanations   & \textbf{0.515} & 0.130          & 0.506          & 0.492          & 1.139          & \textbf{0.569} & 0.114          & \textbf{0.569} & \textbf{0.554} & 1.091          \\
+ ``parallel'' explanations            & 0.409          & 0.134          & 0.404          & 0.407          & 1.150          & 0.472          & 0.123          & 0.483          & 0.472          & 1.118          \\
+ ``parallel'' explicit   explanations & 0.514          & \textbf{0.108} & \textbf{0.507} & \textbf{0.492} & \textbf{1.074} & 0.558          & \textbf{0.092} & 0.558          & 0.542          & \textbf{1.025} \\
\midrule
$\bm{p}_{\text{sfmax}}$ of Mixtral              & 0.413          & 0.131          & 0.427          & 0.408          & 1.140          & 0.477          & 0.121          & 0.497          & 0.460          & 1.112          \\
+ ``serial'' explanations              & 0.467          & 0.121          & 0.452          & 0.459          & 1.113          & 0.504          & 0.110          & 0.506          & 0.498          & 1.078          \\
+ ``serial'' explicit   explanations   & 0.517          & 0.105          & 0.509          & 0.495          & 1.064          & 0.566          & 0.093          & \textbf{0.568} & 0.549          & 1.026          \\
+ ``parallel'' explanations            & 0.404          & 0.121          & 0.397          & 0.402          & 1.112          & 0.483          & 0.110          & 0.497          & 0.482          & 1.079          \\
+ ``parallel'' explicit   explanations & \textbf{0.528} & \textbf{0.095} & \textbf{0.522} & \textbf{0.502} & \textbf{1.035} & \textbf{0.568} & \textbf{0.082} & 0.567          & \textbf{0.549} & \textbf{0.994} \\
\midrule
\multicolumn{4}{l}{\textit{MJDs from Llama3}}   \\
\midrule 
$\bm{p}_{\text{norm}}$ of Llama3                           & 0.556          & 0.097          & 0.514          & 0.473          & 1.038          & 0.593          & 0.091          & 0.541          & 0.505          & 1.023          \\
+ ``serial'' explanations              & 0.561          & 0.096          & 0.574          & 0.534          & 1.037          & 0.610          & 0.091          & 0.618          & 0.580          & 1.020          \\
+ ``serial'' explicit   explanations   & 0.571          & \textbf{0.091} & 0.578          & 0.536          & \textbf{1.022} & 0.630          & \textbf{0.085} & 0.634          & 0.589          & \textbf{1.003} \\
+ ``parallel'' explanations            & 0.561          & 0.098          & 0.573          & 0.531          & 1.041          & 0.634          & 0.093          & 0.636          & 0.594          & 1.026          \\
+ ``parallel'' explicit   explanations & \textbf{0.572} & 0.094          & \textbf{0.582} & \textbf{0.539} & 1.030          & \textbf{0.634} & 0.089          & \textbf{0.639} & \textbf{0.596} & 1.014          \\
\midrule
$\bm{p}_{\text{sfmax}}$ of Llama3             & 0.574          & 0.091          & 0.528          & 0.487          & 1.023          & 0.601          & 0.085          & 0.546          & 0.514          & 1.005          \\
+ ``serial'' explanations              & 0.555          & 0.091          & 0.567          & 0.532          & 1.021          & 0.617          & 0.082          & 0.626          & 0.588          & 0.996          \\
+ ``serial'' explicit   explanations   & \textbf{0.577} & \textbf{0.086} & \textbf{0.585} & \textbf{0.544} & \textbf{1.008} & \textbf{0.641} & \textbf{0.077} & \textbf{0.646} & 0.604          & \textbf{0.981} \\
+ ``parallel'' explanations            & 0.572          & 0.092          & 0.584          & 0.541          & 1.024          & 0.638          & 0.085          & 0.643          & 0.605          & 1.004          \\
+ ``parallel'' explicit   explanations & 0.572          & 0.088          & 0.581          & 0.540          & 1.014          & 0.639          & 0.081          & 0.645          & \textbf{0.605} & 0.993         
\\
\bottomrule
\end{tabular}}
\caption{All results for fine-tuning comparison on Chaos NLI dev set.}\label{tab:main-ft-eval}
\end{table*}

\begin{table*}[htbp]
\centering
\resizebox{\textwidth}{!}{
\begin{tabular}{lccccc|ccccc}
\toprule
\multicolumn{1}{c}{\multirow{2}{*}{\textbf{Distributions}}}  & \multicolumn{5}{c|}{\textbf{BERT FT}} & \multicolumn{5}{c}{\textbf{RoBERTa FT}}\\ \cmidrule(lr){2-11} 
\multicolumn{1}{c}{}                                                   & \textbf{ACC. $\uparrow$} & \textbf{KL $\downarrow$} & \textbf{Weighted F1 $\uparrow$} & \textbf{Macro F1 $\uparrow$}  &  \textbf{CE Loss $\downarrow$}     & \textbf{ACC. $\uparrow$} & \textbf{KL $\downarrow$} & \textbf{Weighted F1 $\uparrow$} & \textbf{Macro F1 $\uparrow$}  &  \textbf{CE Loss $\downarrow$}     \\
\midrule
\multicolumn{10}{l}{\textit{Baseline}}   \\
\midrule
Chaos NLI-M train set                  & \textbf{0.645}          & \textbf{0.077}          & \textbf{0.646}          & \textbf{0.617}          & \textbf{0.974}          & \textbf{0.650}          & \textbf{0.067}          & \textbf{0.650}          & \textbf{0.630}          & \textbf{0.943}          \\
MNLI single label                      & 0.585          & 0.704          & 0.589          & 0.573          & 2.855          & 0.599          & 0.867          & 0.603          & 0.594          & 3.344          \\
MNLI distributions                     & 0.547          & 0.102          & 0.543          & 0.539          & 1.048          & 0.599          & 0.096          & 0.604          & 0.594          & 1.029          \\
VariErr distributions                  & 0.566          & 0.186          & 0.559          & 0.548          & 1.299          & 0.590          & 0.197          & 0.589          & 0.569          & 1.333          \\
\midrule
\multicolumn{4}{l}{\textit{MJDs from Mixtral}}   \\
\midrule           
$\bm{p}_{\text{norm}}$ of Mixtral                    & 0.417          & 0.133          & 0.422          & 0.416          & 1.142          & 0.459          & 0.127          & 0.466          & 0.453          & 1.123          \\
+ ``serial'' explanations              & 0.479          & 0.141          & 0.454          & 0.466          & 1.166          & 0.517          & 0.128          & 0.514          & 0.514          & 1.126          \\
+ ``serial'' explicit   explanations   & 0.523          & 0.130          & 0.511          & 0.509          & 1.132          & \textbf{0.572} & 0.122          & \textbf{0.572} & \textbf{0.566} & 1.107          \\
+ ``parallel'' explanations            & 0.436          & 0.131          & 0.428          & 0.434          & 1.136          & 0.498          & 0.122          & 0.502          & 0.499          & 1.109          \\
+ ``parallel'' explicit   explanations & \textbf{0.526} & \textbf{0.108} & \textbf{0.514} & \textbf{0.510} & \textbf{1.065} & 0.568          & \textbf{0.098} & 0.565          & 0.559          & \textbf{1.037} \\
\midrule
$\bm{p}_{\text{sfmax}}$ of Mixtral                         & 0.423          & 0.129          & 0.432          & 0.421          & 1.130          & 0.464          & 0.125          & 0.472          & 0.457          & 1.118          \\
+ ``serial'' explanations              & 0.485          & 0.118          & 0.462          & 0.473          & 1.096          & 0.526          & 0.109          & 0.525          & 0.522          & 1.069          \\
+ ``serial'' explicit   explanations   & \textbf{0.531} & 0.105          & \textbf{0.520} & \textbf{0.517} & 1.057          & 0.574          & 0.098          & 0.573          & 0.567          & 1.036          \\
+ ``parallel'' explanations            & 0.436          & 0.119          & 0.429          & 0.433          & 1.098          & 0.501          & 0.111          & 0.505          & 0.501          & 1.074          \\
+ ``parallel'' explicit   explanations & 0.528          & \textbf{0.095} & 0.517          & 0.510          & \textbf{1.026} & \textbf{0.577} & \textbf{0.087} & \textbf{0.576} & \textbf{0.567} & \textbf{1.003} \\
\midrule
\multicolumn{4}{l}{\textit{MJDs from Llama3}}   \\
\midrule               
$\bm{p}_{\text{norm}}$ of Llama3                 & 0.561          & 0.098          & 0.526          & 0.506          & 1.036          & 0.583          & 0.094          & 0.528          & 0.513          & 1.025          \\
+ ``serial'' explanations              & 0.568          & 0.097          & 0.574          & 0.555          & 1.033          & 0.596          & 0.093          & 0.601          & 0.585          & 1.022          \\
+ ``serial'' explicit   explanations   & 0.571          & \textbf{0.092} & 0.574          & 0.550          & \textbf{1.018} & 0.595          & \textbf{0.088} & 0.598          & 0.581          & \textbf{1.006} \\
+ ``parallel'' explanations            & 0.572          & 0.098          & 0.582          & 0.560          & 1.038          & 0.595          & 0.095          & 0.598          & 0.579          & 1.028          \\
+ ``parallel'' explicit   explanations & \textbf{0.580} & 0.095          & \textbf{0.586} & \textbf{0.560} & 1.026          & \textbf{0.615} & 0.091          & \textbf{0.620} & \textbf{0.598} & 1.016          \\
\midrule
$\bm{p}_{\text{sfmax}}$ of Llama3                    & 0.564          & 0.093          & 0.524          & 0.507          & 1.021          & 0.595          & 0.089          & 0.535          & 0.524          & 1.009          \\
+ ``serial'' explanations              & 0.569          & 0.091          & 0.576          & 0.556          & 1.016          & 0.603          & 0.086          & 0.608          & 0.591          & 1.000          \\
+ ``serial'' explicit   explanations   & 0.566          & \textbf{0.087} & 0.568          & 0.548          & \textbf{1.004} & 0.607          & \textbf{0.081} & 0.610          & 0.591          & \textbf{0.987} \\
+ ``parallel'' explanations            & \textbf{0.574}          & 0.093          & \textbf{0.583} & \textbf{0.560} & 1.020          & 0.606          & 0.089          & 0.611          & 0.590          & 1.008          \\
+ ``parallel'' explicit   explanations & 0.574 & 0.089          & 0.578          & 0.551          & 1.010          & \textbf{0.617} & 0.085          & \textbf{0.621} & \textbf{0.601} & 0.996         
\\
\bottomrule
\end{tabular}}
\caption{All results for fine-tuning comparison on Chaos NLI test set.}\label{tab:main-ft-test}
\end{table*}

\section{Detailed Result of Ablation Study}
\label{app:ablation}

All results of the ablation study (``serial''/``parallel'') were listed in Table~\ref{tab:ablation}, which were depicted by Figure~\ref{fig:ablation} in \S\ref{subsec:experiment-distribution-comparison}.
To pursue maximizing the evaluation metrics, we can indeed make greater sacrifices in computational resources, especially since Llama3 has a strong capability to process all explanations at once. However, we have considered the possible sequence bias of explanations. To eliminate these biases we consider a shuffling setup, where the computational resource consumption will increase significantly as the number of explanations increases. Therefore, presenting both parallel and serial results is valuable.

\section{Details of Fine-tuning Comparison}
\label{app:detail}

Here we provided more details regarding the results of \emph{fine-tuning} comparison, as listed in Table~\ref{tab:main-ft-eval} and ~\ref{tab:main-ft-test}, which were the complete versions of Table~\ref{tab:main-ft}. Metrics of KL, CE Loss, Accuracy, Weighted F1 and Macro F1 were included. The same pattern was observed as illustrated in \S\ref{subsec:results-finetuning-comparisons}.

\section{Detailed Visualizations}
\label{app:visual}

\begin{figure}[htbp]
	\centering
	\begin{subfigure}{\linewidth}
		\centering
		\includegraphics[width=\linewidth]{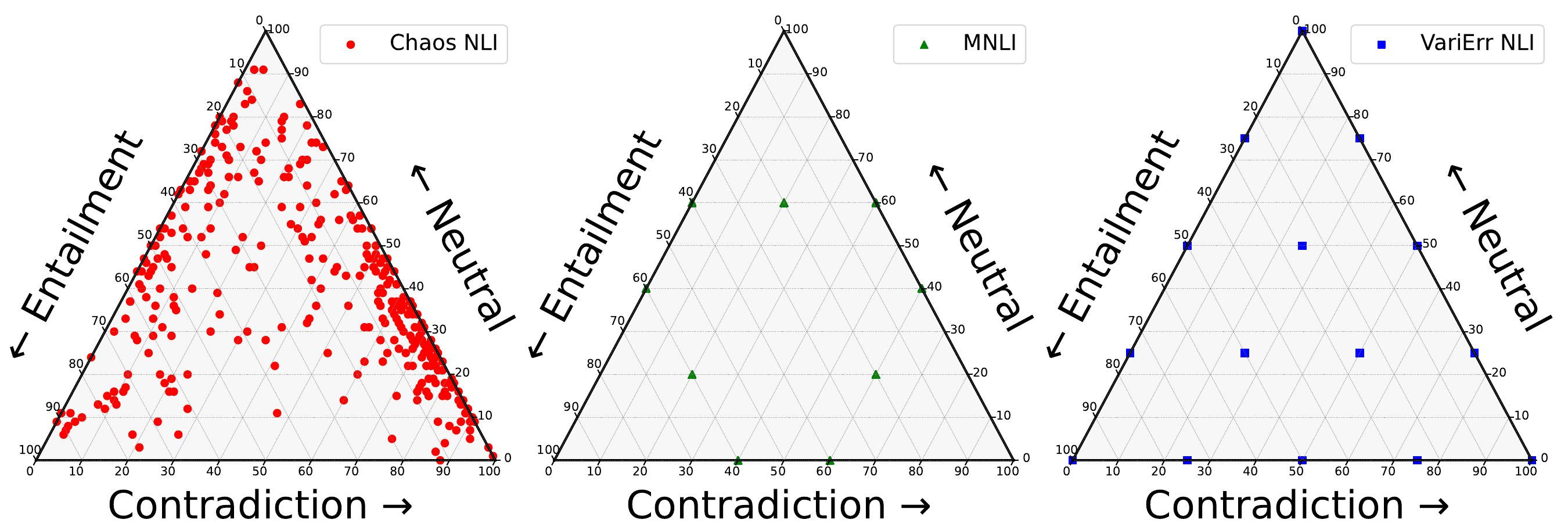}
		\caption{Distributions of datasets.}
		\label{fig:visual-a}
	\end{subfigure}

	\centering
	\begin{subfigure}{\linewidth}
		\centering
		\includegraphics[width=\linewidth]{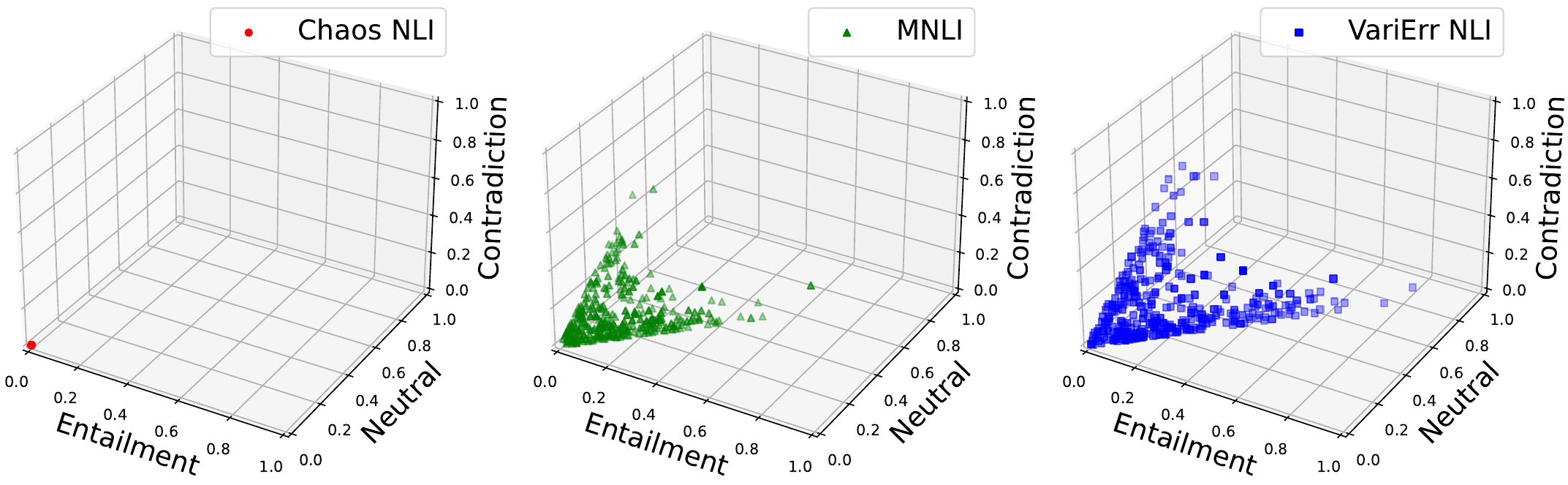}
		\caption{Absolute errors of Datasets.}
		\label{fig:visual-b}
	\end{subfigure}

	\centering
	\begin{subfigure}{\linewidth}
		\centering
		\includegraphics[width=\linewidth]{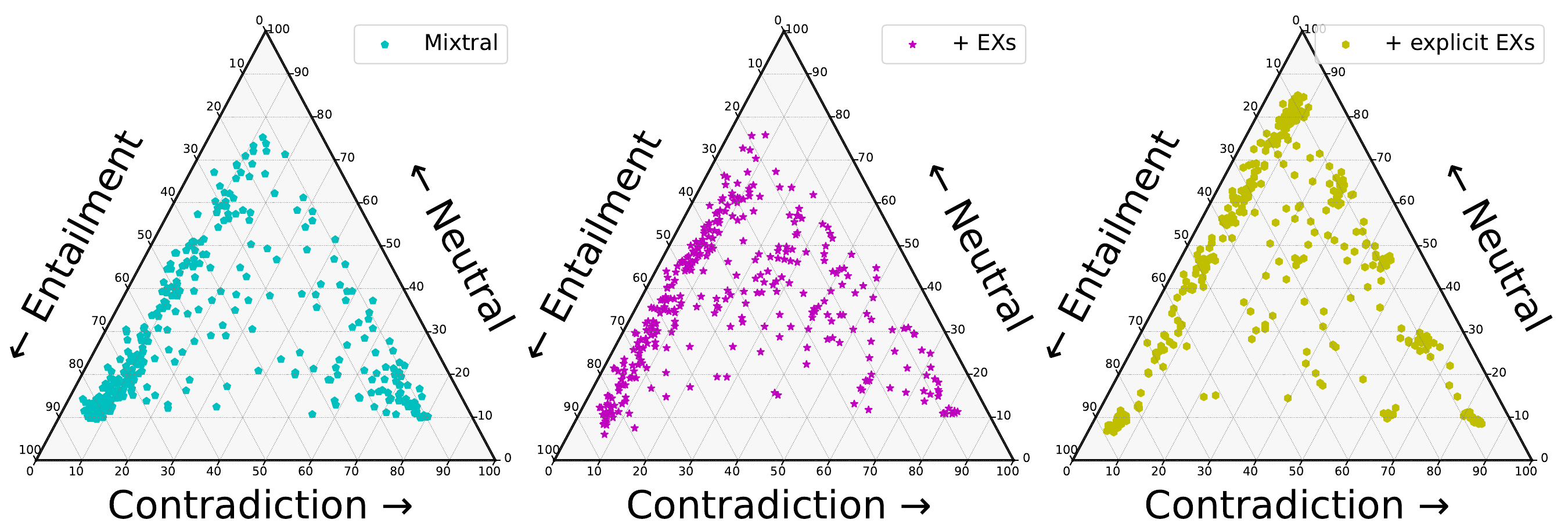}
		\caption{$\bm{p}_{\text{norm}}$ of Mixtral.}
		\label{fig:visual-c}
	\end{subfigure}

        \centering
	\begin{subfigure}{\linewidth}
		\centering
		\includegraphics[width=\linewidth]{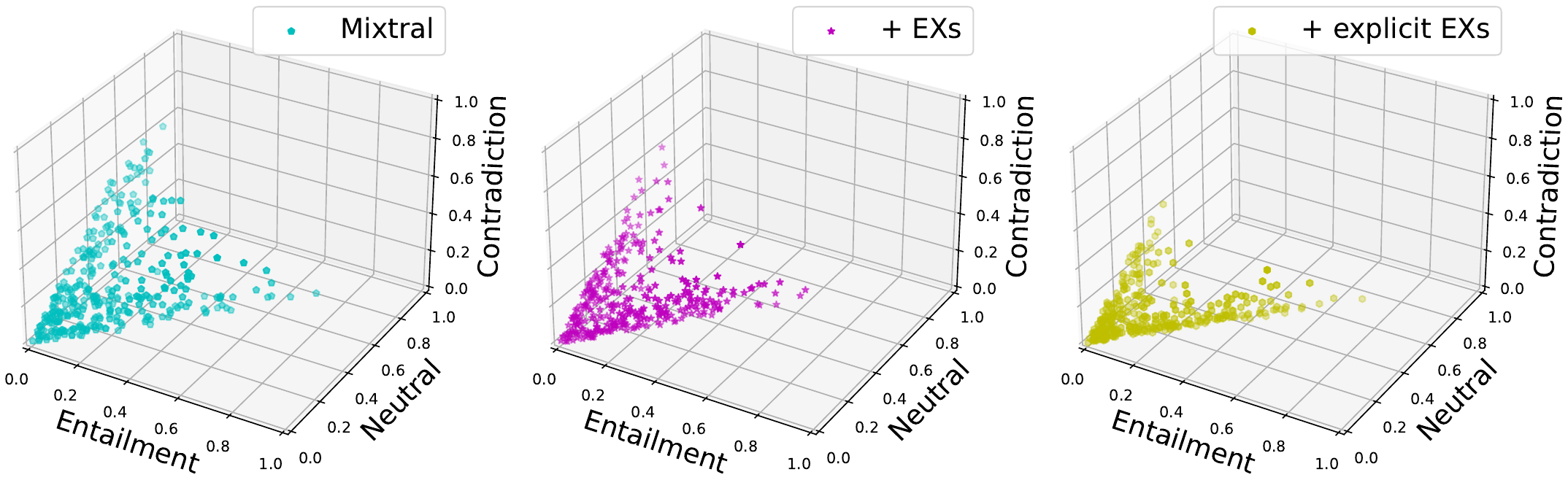}
		\caption{Absolute errors of Mixtral.}
		\label{fig:visual-d}
	\end{subfigure}
 
	\centering
	\begin{subfigure}{\linewidth}
		\centering
		\includegraphics[width=\linewidth]{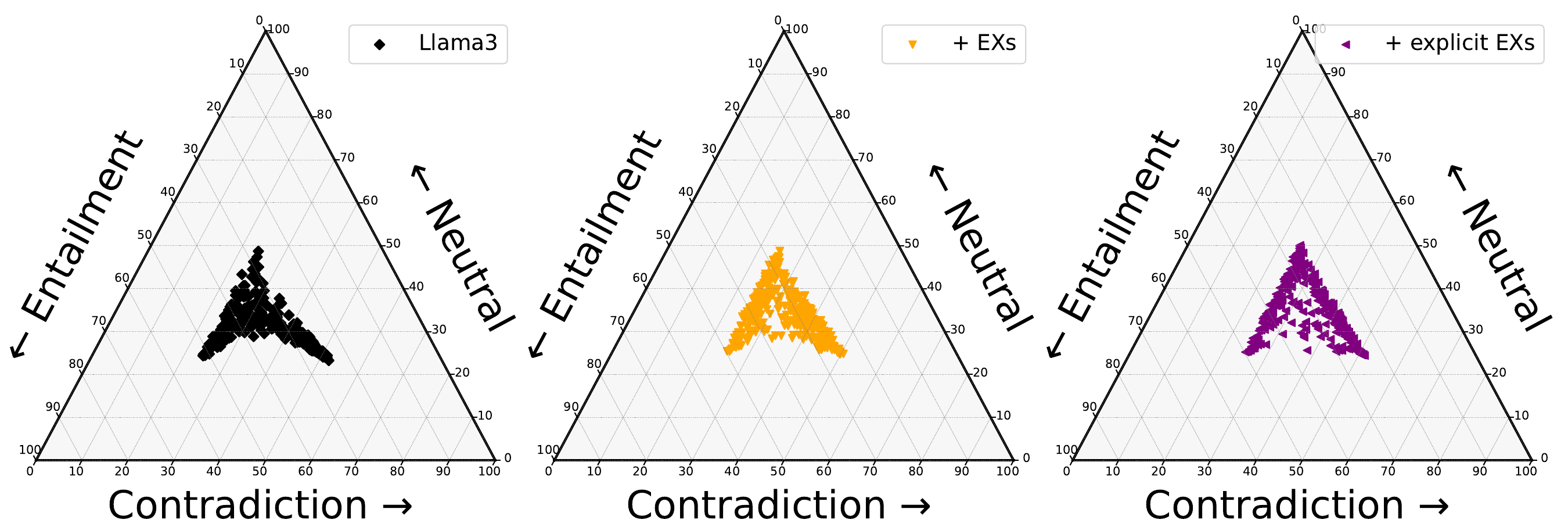}
		\caption{$\bm{p}_{\text{norm}}$ of Llama3.}
		\label{fig:visual-e}
	\end{subfigure}

	\centering
	\begin{subfigure}{\linewidth}
		\centering
		\includegraphics[width=\linewidth]{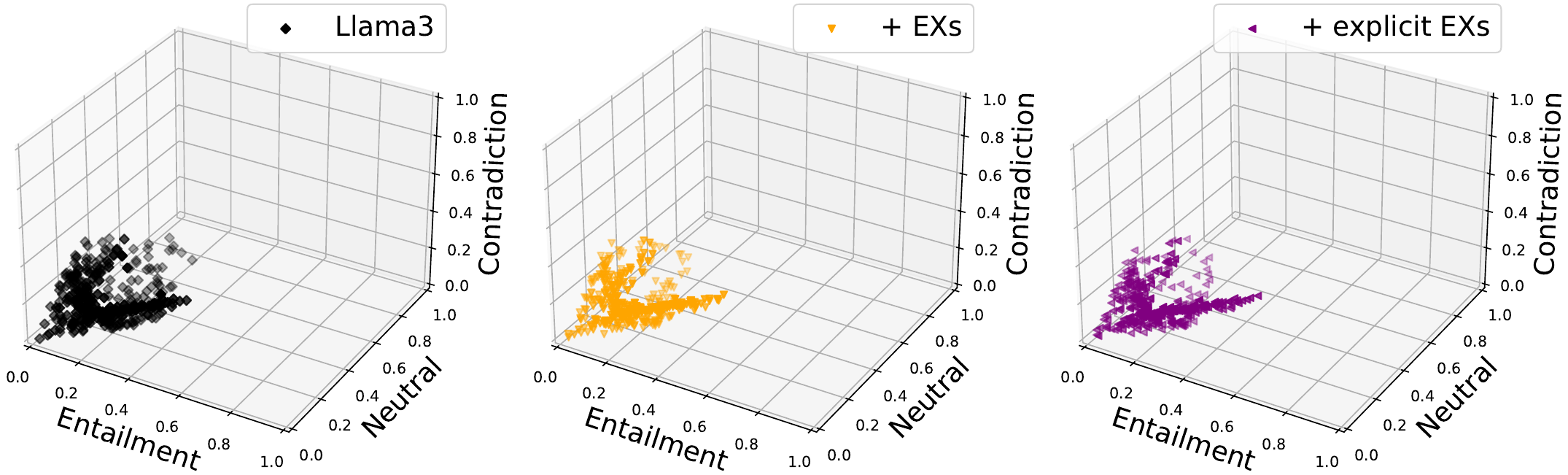}
		\caption{Absolute errors of Llama3.}
		\label{fig:visual-f}
	\end{subfigure}
	\caption{Visualization of distributions and absolute errors in ternary plot and 3D Cartesian coordinate system. Each point represents one sample from the valid overlapped 341 instances.}
	\label{fig:visual}
\end{figure}

As shown in Figure~\ref{fig:visual}, we plotted the distributions and the absolute errors from Chaos NLI HLV to provide more details. When (explicit) explanations were added, the absolute errors of MJDs became smaller, meaning they got closer to HJD.

\section{Data Leakage for LLMs}
\label{app:leakage}

The lack of complete pre-training details on Large Language Models (LLMs) raises the potential risk of data contamination and the skepticism on LLM performance evaluation \citep{balloccu-etal-2024-leak, behnamghader2024llm2vec}.
To the best of our knowledge, there lacks concrete evidence showing either 
Mixtral-8x7b-Instruct-v0.1 \cite{DBLP:journals/corr/abs-2401-04088} 
or Llama3-Chat-70b \cite{llama3} being contaminated by any of these tested datasets, MNLI \citep{DBLP:conf/naacl/WilliamsNB18},
VariErr NLI \citep{DBLP:conf/nips/Wei0SBIXCLZ22},
and Chaos NLI \citep{DBLP:conf/emnlp/NieZB20}. 
For the VariErr NLI dataset that provides explanations, its release date is later than that of Mixtral and Llama3, so it does not pose a leakage risk. 
Moreover, closely related to our datasets,  \citet{ahuja2024megaverse} conduct the Black Box contamination test \citep{oren2024proving} and show that XNLI  \citep{conneau-etal-2018-xnli}, a sub-sampled multilingual extension dataset of MNLI, did not leak data to either the Llama-2-7B \citep{touvron2023llama} or the Mistral-7B-Instruct \citep{jiang2023mistral} model. 
Since Mixtral or Llama have only released their weights, we do not have access to their training corpora. Therefore, we can only offer our perspectives and discussions on data leakage.

For the MNLI training part,
to familiarize BERT and RoBERTa with the NLI task, we first fine-tuned the models on MNLI training set and then further fine-tuned them on our samples, which contain both HJDs and MJDs.
Please note that in our fine-tuning comparison, the "gold" labels are derived from the distribution of 100 annotations in ChaosNLI, not the single labels from MNLI. Besides, the corpora we evaluated, 341 ChaosNLI training set and 629 dev/test set, were all extracted from MNLI-matched development set \cite{DBLP:conf/emnlp/NieZB20}, which did not overlap with MNLI training set used for MNLI fine-tuning. Therefore, there is no risk of data leakage in this process.

\section{Discussion for Temperature in Softmax}
\label{app:discuss_t}

Equation~\ref{equation2} is a standard softmax function used to convert the logits obtained from the LLM into a label probability distribution. \citet{zhou-etal-2022-distributed} explores various methods for obtaining probability distributions, including Monte Carlo Dropout, among others. Since these methods are not significantly different from the conventional softmax, and exploring which normalization method is better is not the focus of this paper, we have directly adopted the most commonly used non-parametric normalization and parametric softmax approach. In the softmax function, the parameter $\tau$ is often referred to as the temperature coefficient, which smooths the probability distribution (a smaller $\tau$ makes the label distribution sharper, while a larger $\tau$ makes it smoother). 
Typically, for converting LLM logits into a probability distribution, $\tau$ should be set to a moderately sized value to prevent excessively small $\tau$ values from sharpening the probability distribution, thereby potentially losing the HLV information we aim to obtain.

Thus, during the \emph{distribution comparison}, we adopted $\tau=20$ to obtain $\bm{p}_{\text{sfmax}}$ that can be compared with $\bm{p}_{\text{norm}}$. In fact, for the metrics in Table~\ref{tab:main-di}, the results of $\bm{p}_{\text{sfmax}}$ improve initially as $\tau$ increases from 0, and then decline, as shown in our toy trials. However, we did not specifically adjust the value of 
$\tau$ because it is not relevant to the focus of our study. $\bm{p}_{\text{norm}}$ does not have any parameters and still yields the desired results.

\begin{figure}[htbp]
	\centering
 	\begin{subfigure}{\linewidth}
		\centering
		\includegraphics[width=\linewidth]{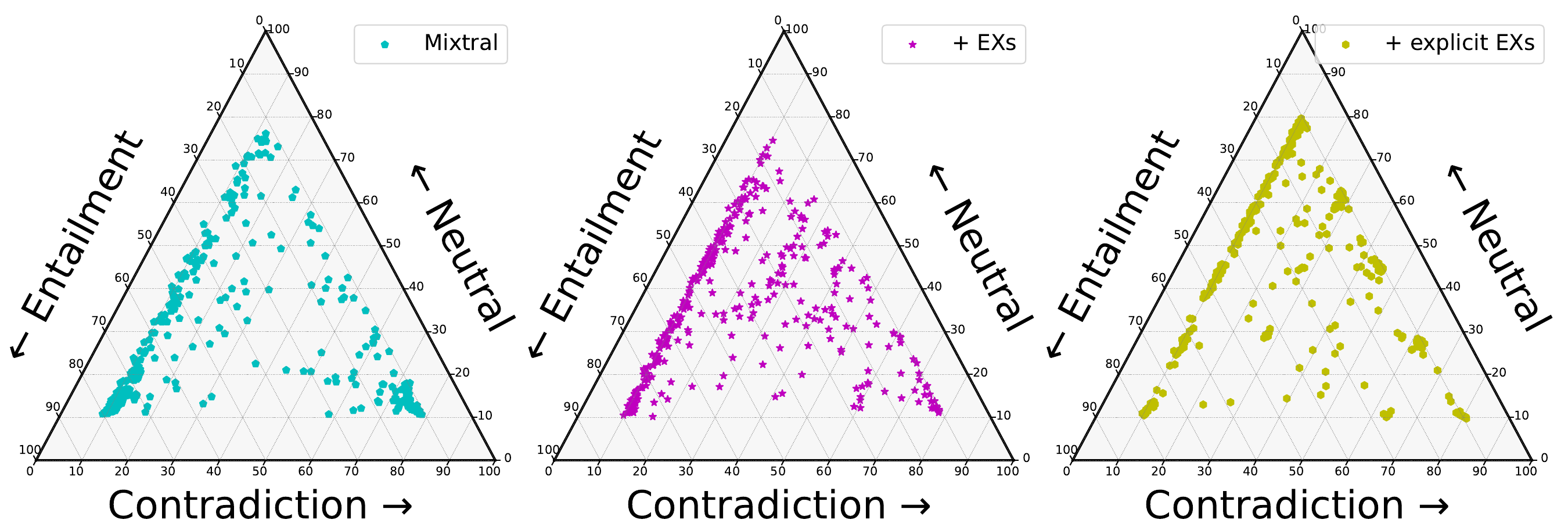}
		\caption{$\bm{p}_{\text{sfmax}}$ of Mixtral ($\tau = 20$).}
		\label{fig:discuss_t20_mix}
	\end{subfigure}
 
	\centering
	\begin{subfigure}{\linewidth}
		\centering
		\includegraphics[width=\linewidth]{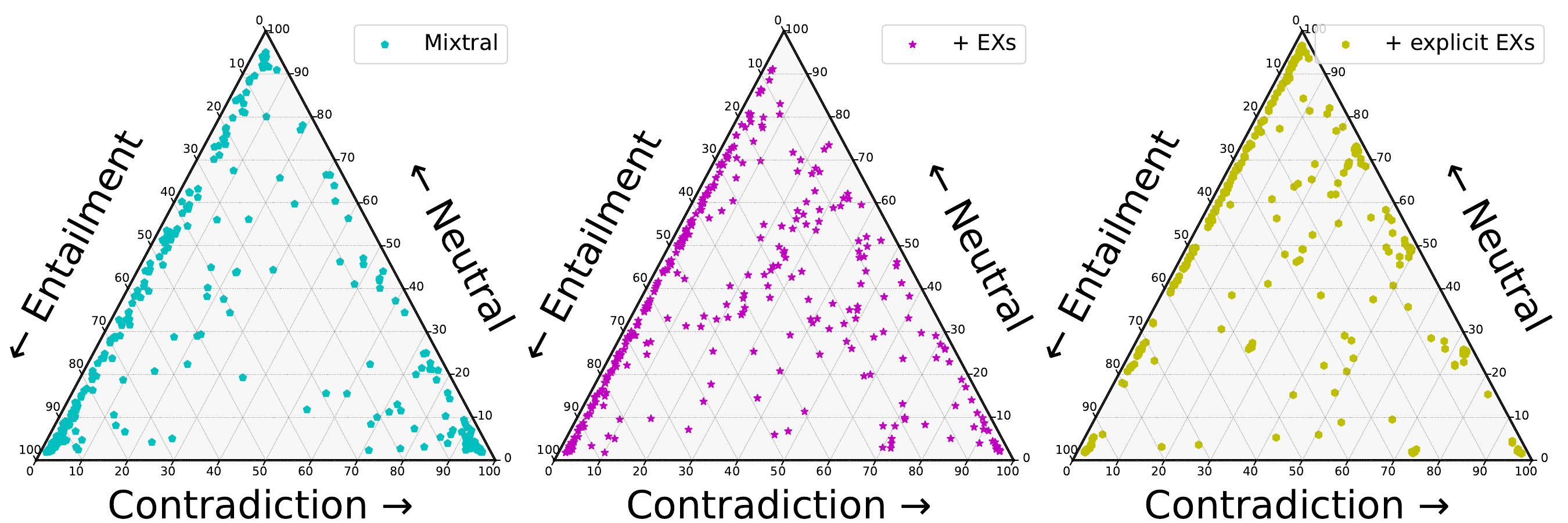}
		\caption{$\bm{p}_{\text{sfmax}}$ of Mixtral ($\tau = 10$).}
		\label{fig:discuss_t10_mix}
	\end{subfigure}

	\centering
	\begin{subfigure}{\linewidth}
		\centering
		\includegraphics[width=\linewidth]{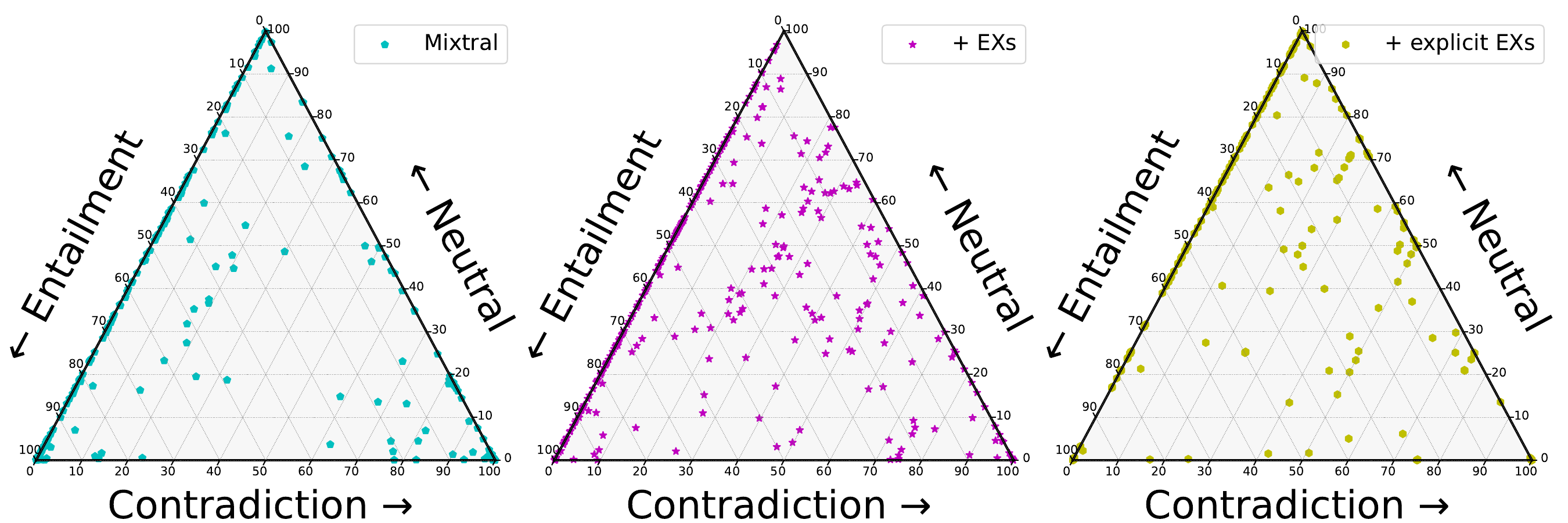}
		\caption{$\bm{p}_{\text{sfmax}}$ of Mixtral ($\tau = 5$).}
		\label{fig:discuss_t5_mix}
	\end{subfigure}

        \centering
	\begin{subfigure}{\linewidth}
		\centering
		\includegraphics[width=\linewidth]{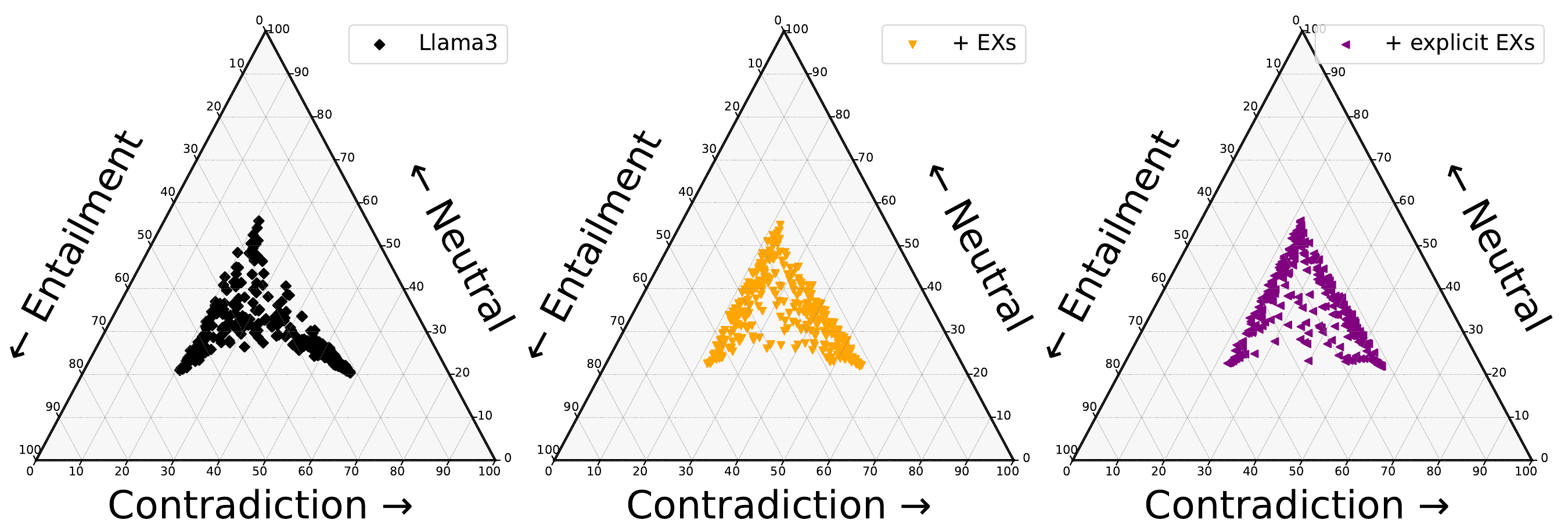}
		\caption{$\bm{p}_{\text{sfmax}}$ of Llama3 ($\tau = 20$).}
		\label{fig:discuss_t20}
	\end{subfigure}
 
	\centering
	\begin{subfigure}{\linewidth}
		\centering
		\includegraphics[width=\linewidth]{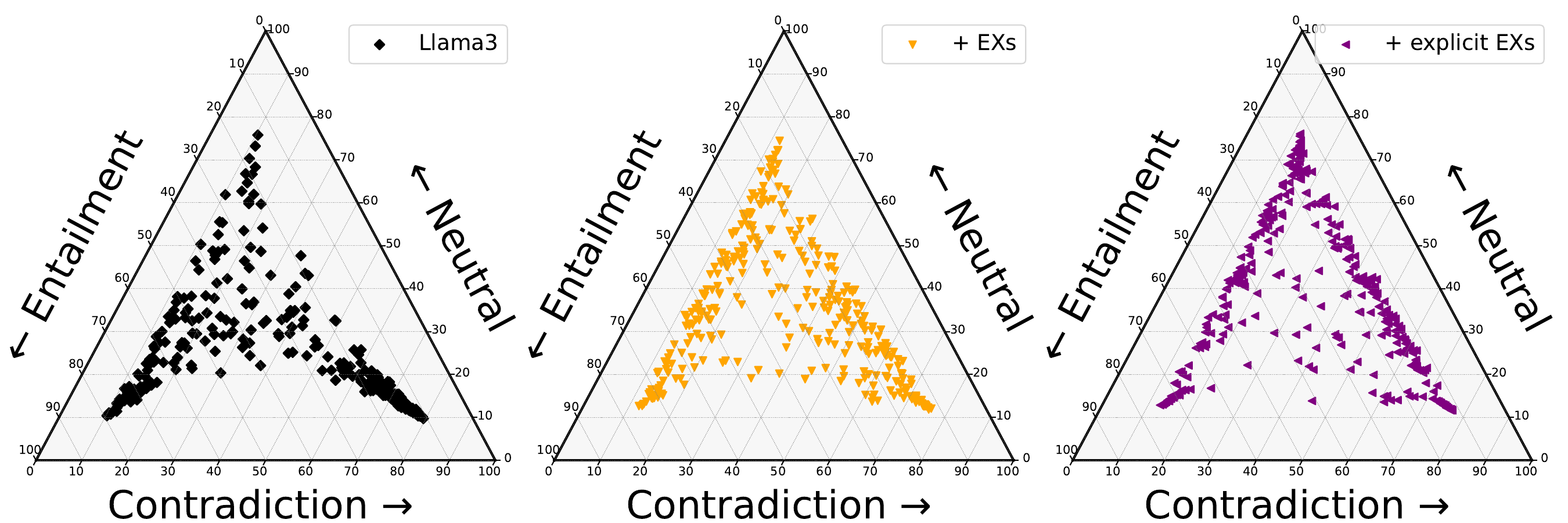}
		\caption{$\bm{p}_{\text{sfmax}}$ of Llama3 ($\tau = 10$).}
		\label{fig:discuss_t10}
	\end{subfigure}

	\centering
	\begin{subfigure}{\linewidth}
		\centering
		\includegraphics[width=\linewidth]{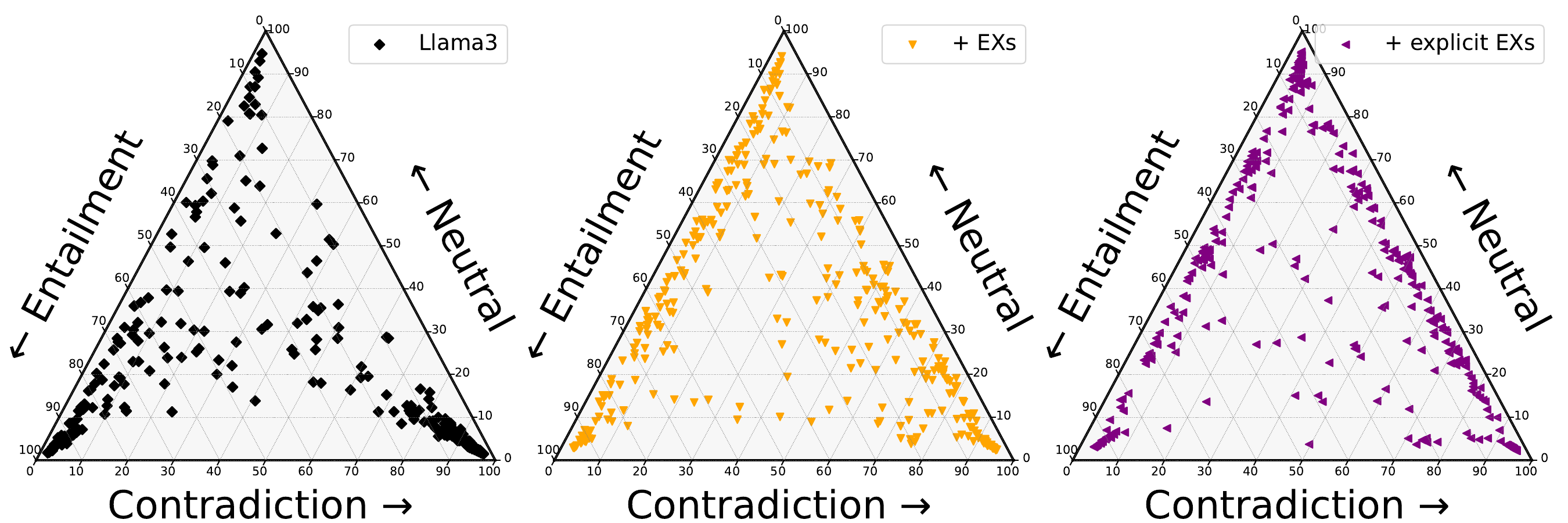}
		\caption{$\bm{p}_{\text{sfmax}}$ of Llama3 ($\tau = 5$).}
		\label{fig:discuss_t5}
	\end{subfigure}
	\caption{Visualization of distributions in ternary plot for $\tau$ discussion. Through amplification, the bad becomes worse, and the good becomes better.}
	\label{fig:discuss_t}
\end{figure}

During our in-depth investigation at \S\ref{subsec:analysis-visualization}, we discovered that the value of 
$\tau$ has an amplifying effect on the distribution shape within a certain range (not a true proportional amplification, but very similar within that range). Given our focus on exploring possible amplification patterns in the Llama3 MJDs, we adjusted $\tau$ from 20 to 10 to 5. This adjustment allowed us to better observe the intrinsic shape of the Llama3 MJD and investigate its correlation with Chaos NLI HJD. All the ``scaled'' MJDs are shown in Figure~\ref{fig:discuss_t}, including distributions of Mixtral and Llama3 under different settings. Also, we conducted \emph{fine-tuning comparison}, same as \S\ref{subsec:results-finetuning-comparisons}, and listed all results in Table~\ref{tab:softmax-t}, for further analysis.

The results show that as $\tau$ decreases, the distribution becomes sharper, but this does not result in a linear change in metrics such as F1, KL divergence, and CE Loss. Since this parameter is highly dependent on the dataset and model, we did not spend extensive computational resources on a broader exploration of $\tau$ values.

\begin{table*}[htbp]
\centering
\resizebox{\textwidth}{!}{
\begin{tabular}{lccc|ccc}
\toprule
\multicolumn{1}{c}{\multirow{2}{*}{\textbf{Distributions}}}  & \multicolumn{3}{c|}{\textbf{BERT FT (dev / test)}} & \multicolumn{3}{c}{\textbf{RoBERTa FT (dev / test)}}\\ \cmidrule(lr){2-7} 
\multicolumn{1}{c}{}                         & \textbf{Weighted F1 $\uparrow$}                      & \textbf{KL $\downarrow$} &  \textbf{CE Loss $\downarrow$}   & \textbf{Weighted F1 $\uparrow$}  & \textbf{KL $\downarrow$}  &  \textbf{CE Loss $\downarrow$}     \\
\midrule
\multicolumn{7}{l}{\textit{Baseline}}   \\
\midrule
Chaos NLI train set                  & \textbf{0.626} / \textbf{0.646} & \textbf{0.074} / \textbf{0.077} & \textbf{0.972} / \textbf{0.974} & \textbf{0.699} / \textbf{0.650} & \textbf{0.061} / \textbf{0.067} & \textbf{0.932} / \textbf{0.943} \\
MNLI single label                      & 0.561          / 0.589          & 0.665          / 0.704          & 2.743          / 2.855          & 0.635          / 0.603          & 0.844          / 0.867          & 3.281          / 3.344          \\
MNLI distributions                     & 0.546          / 0.543          & 0.099          / 0.102          & 1.046          / 1.048          & 0.613          / 0.604          & 0.100          / 0.096          & 1.047          / 1.029          \\
VariErr distributions                  & 0.557          / 0.559          & 0.179          / 0.186          & 1.286          / 1.299          & 0.617          / 0.589          & 0.174          / 0.197          & 1.269          / 1.333          \\
\midrule
\multicolumn{7}{l}{\textit{MJDs from Mixtral}}   \\
\midrule
$\bm{p}_{\text{norm}}$ of Mixtral                                & 0.416          / 0.422          & 0.134          / 0.133          & 1.152          / 1.142          & 0.486          / 0.466          & 0.123          / 0.127          & 1.118          / 1.123          \\
+ ``serial'' explanations              & 0.443          / 0.454          & 0.145          / 0.141          & 1.183          / 1.166          & 0.509          / 0.514          & 0.128          / 0.128          & 1.132          / 1.126          \\
+ ``serial'' explicit   explanations   & 0.506          / 0.511          & 0.130          /0.130          & 1.139          / 1.132          & \textbf{0.569} / \textbf{0.572} & 0.114          / 0.122          & 1.091          / 1.107          \\
+ ``parallel'' explanations            & 0.404          / 0.428          & 0.134          / 0.131          & 1.150          / 1.136          & 0.483          / 0.502          & 0.123          / 0.122          & 1.118          / 1.109          \\
+ ``parallel'' explicit   explanations & \textbf{0.507} / \textbf{0.514} & \textbf{0.108} / \textbf{0.108} & \textbf{1.074} / \textbf{1.065} & 0.558          / 0.565          & \textbf{0.092} / \textbf{0.098} & \textbf{1.025} / \textbf{1.037} \\
\midrule
$\bm{p}_{\text{sfmax}}$ of Mixtral (\(\tau = 20\))                               & 0.427          / 0.432          & 0.131          / 0.129          & 1.140          / 1.130          & 0.497          / 0.472          & 0.121          / 0.125          & 1.112          / 1.118          \\
+ ``serial'' explanations              & 0.452          / 0.462          & 0.121          / 0.118          & 1.113          / 1.096          & 0.506          / 0.525          & 0.110          / 0.109          & 1.078          / 1.069          \\
+ ``serial'' explicit   explanations   & 0.509          / \textbf{0.520} & 0.105          / 0.105          & 1.064          / 1.057          & \textbf{0.568} / 0.573          & 0.093          / 0.098          & 1.026          / 1.036          \\
+ ``parallel'' explanations            & 0.397          / 0.429          & 0.121          / 0.119          & 1.112          / 1.098          & 0.497          / 0.505          & 0.110          / 0.111          & 1.079          / 1.074          \\
+ ``parallel'' explicit   explanations & \textbf{0.522} / 0.517          & \textbf{0.095} / \textbf{0.095} & \textbf{1.035} / \textbf{1.026} & 0.567          / \textbf{0.576} & \textbf{0.082} / \textbf{0.087} & \textbf{0.994} / \textbf{1.003} \\
\midrule
$\bm{p}_{\text{sfmax}}$ of Mixtral (\(\tau = 10\)) &
 0.445          / 0.435          & 0.210          / 0.214          & 1.380          / 1.384          & 0.487          / 0.492          & 0.207          / 0.209          & 1.369          / 1.370          \\
+ ``serial'' explanations    & 0.457          / 0.461          & 0.198          / 0.195          & 1.344          / 1.328          & 0.522          / 0.533          & 0.181          / 0.182          & 1.290          / 1.288          \\
+ ``serial'' explicit   explanations& \textbf{0.522} / \textbf{0.532} & 0.184          / 0.188          & 1.302          / 1.306          & \textbf{0.599} / \textbf{0.584} & 0.176          / 0.191          & 1.275          / 1.314          \\
+ ``parallel'' explanations &0.426          / 0.432          & 0.175          / 0.176          & 1.275          /  1.269          & 0.518          / 0.514          & 0.157          / 0.158          & 1.221          / 1.217          \\
+ ``parallel'' explicit   explanations &0.521          / 0.524          & \textbf{0.143} / \textbf{0.145} & \textbf{1.177} / \textbf{1.176} & 0.576          / 0.567          & \textbf{0.126} / \textbf{0.140} & \textbf{1.128} / \textbf{1.162} \\
\midrule
$\bm{p}_{\text{sfmax}}$ of Mixtral (\(\tau = 5\)) & 0.453          / 0.439          & 0.330          / 0.338          & 1.737          / 1.757          & 0.559          / 0.542          & 0.143          / 0.150          & 1.177          / 1.192          \\
+ ``serial'' explanations   & 0.452          / 0.464          & 0.287          / 0.287          & 1.610          / 1.604          & 0.573          / 0.581          & 0.113          / 0.116          & 1.088          / 1.091          \\
+ ``serial'' explicit   explanations & \textbf{0.525} / \textbf{0.530} & 0.295          / 0.308          & 1.634          / 1.665          & \textbf{0.593} / 0.588          & 0.131          / 0.138          & 1.142          / 1.156          \\
+ ``parallel'' explanations & 0.435          / 0.430          & 0.217          / 0.221          & 1.398          / 1.406          & 0.593          / \textbf{0.599} & \textbf{0.095} / \textbf{0.098} & \textbf{1.034} / \textbf{1.037} \\
+ ``parallel'' explicit   explanations & 0.520          / 0.527          & \textbf{0.180} / \textbf{0.184} & \textbf{1.289} / \textbf{1.294} & 0.589          / 0.574          & 0.098          / 0.101          & 1.043          / 1.045          \\
\midrule
\multicolumn{7}{l}{\textit{MJDs from Llama3}}   \\
\midrule
$\bm{p}_{\text{norm}}$ of Llama3                                  & 0.514          / 0.526          & 0.097          / 0.098          & 1.038          / 1.036          & 0.541          / 0.528          & 0.091          / 0.094          & 1.023          / 1.025          \\
+ ``serial'' explanations              & 0.574          / 0.574          & 0.096          / 0.097          & 1.037          / 1.033          & 0.618          / 0.601          & 0.091          / 0.093          & 1.020          / 1.022          \\
+ ``serial'' explicit   explanations   & 0.578          / 0.574          & \textbf{0.091} / \textbf{0.092} & \textbf{1.022} / \textbf{1.018} & 0.634          / 0.598          & \textbf{0.085} / \textbf{0.088} & \textbf{1.003} / \textbf{1.006} \\
+ ``parallel'' explanations            & 0.573          / 0.582          & 0.098          / 0.098          & 1.041          / 1.038          & 0.636          / 0.598          & 0.093          / 0.095          & 1.026          / 1.028          \\
+ ``parallel'' explicit   explanations & \textbf{0.582} / \textbf{0.586} & 0.094          / 0.095          & 1.030          / 1.026          & \textbf{0.639} / \textbf{0.620} & 0.089          / 0.091          & 1.014          / 1.016          \\
\midrule
$\bm{p}_{\text{sfmax}}$ of Llama3  (\(\tau = 20\))                                & 0.528          / 0.524          & 0.091          / 0.093          & 1.023          / 1.021          & 0.546          / 0.535          & 0.085          / 0.089          & 1.005          / 1.009          \\
+ ``serial'' explanations              & 0.567          / 0.576          & 0.091          / 0.091          & 1.021          / 1.016          & 0.626          / 0.608          & 0.082          / 0.086          & 0.996          / 1.000          \\
+ ``serial'' explicit   explanations   & \textbf{0.585} / 0.568          & \textbf{0.086} / \textbf{0.087} & \textbf{1.008} / \textbf{1.004} & \textbf{0.646} / 0.610          & \textbf{0.077} / \textbf{0.081} & \textbf{0.981} / \textbf{0.987} \\
+ ``parallel'' explanations            & 0.584          / \textbf{0.583} & 0.092          / 0.093          & 1.024          / 1.020          & 0.643          / 0.611          & 0.085          / 0.089          & 1.004          / 1.008          \\
+ ``parallel'' explicit   explanations & 0.581          / 0.578          & 0.088          / 0.089          & 1.014          / 1.010          & 0.645          / \textbf{0.621} & 0.081          / 0.085          & 0.993          / 0.996             
\\
\midrule
$\bm{p}_{\text{sfmax}}$ of Llama3  (\(\tau = 10\))  &
 0.550          / 0.541          & 0.090          / 0.093          & 1.020          / 1.021          & 0.571          / 0.547          & 0.085         / 0.089          & 1.003          / 1.009          \\
+ ``serial'' explanations & 0.571          / 0.582          & 0.084          / 0.085          & 1.000          / 0.996          & 0.644          / 0.619          & 0.072          / 0.078          & 0.964          / 0.976          \\
+ ``serial'' explicit   explanations   &0.582          / \textbf{0.602} & 0.081          / 0.083          & 0.993          / 0.992          & 0.656          / \textbf{0.621} & 0.070          / 0.079          & 0.960          / 0.978          \\
+ ``parallel'' explanations & \textbf{0.588} / 0.588          & 0.083          / 0.084          & 0.998          / 0.995          & 0.649          / 0.612          & 0.071          / 0.077          & 0.962          / 0.973          \\
+ ``parallel'' explicit   explanations & 0.576          / 0.587          & \textbf{0.079} / \textbf{0.080} & \textbf{0.985} / \textbf{0.982} & \textbf{0.661} / 0.616          & \textbf{0.066} / \textbf{0.073} & \textbf{0.946} / \textbf{0.960} \\
\midrule
$\bm{p}_{\text{sfmax}}$ of Llama3  (\(\tau = 5\))   & 0.484          / 0.502          & 0.345          / 0.347          & 1.785          / 1.783          & 0.588          / 0.555          & 0.144          / 0.150          & 1.182          / 1.193          \\
+ ``serial'' explanations& 0.531          / 0.525          & 0.271          / 0.276          & 1.563          / 1.570          & 0.657          / 0.621          & 0.104          / 0.118          & 1.061          / 1.097          \\
+ ``serial'' explicit   explanations  & \textbf{0.599} / \textbf{0.590} & 0.311          / 0.331          & 1.682          / 1.735          & 0.663          / 0.622          & 0.132          / 0.150          & 1.145          / 1.192          \\
+ ``parallel'' explanations & 0.516          / 0.524          & 0.207          / 0.210          & 1.370          / 1.373          & 0.658          / 0.611          & \textbf{0.082} / \textbf{0.093} & \textbf{0.995} / \textbf{1.022} \\
+ ``parallel'' explicit   explanations  & 0.577          / 0.568          & \textbf{0.171} / \textbf{0.191} & \textbf{1.262} / \textbf{1.315} & \textbf{0.672} / \textbf{0.623} & 0.083          / 0.096          & 0.996          / 1.030          \\
\bottomrule
\end{tabular}}
\caption{Results for ``\(\tau\)'' discussion of fine-tuning comparison on Chaos NLI dev/test set.
The KL and Cross-Entropy (CE) Loss reflected the distance between distributions, whereas Weighted F1 reflected the capability to handle NLI problems. When $\tau$ gets smaller the F1 of Llama3 will improve. However, the change of KL and CE Loss is not linear. It seems we could find a balance point for both performance between single label evaluation and distribution evaluation by searching $\tau$, which could be a future study. }\label{tab:softmax-t}
\end{table*}

\section{All Results for Distance Correlation}
\label{app:new_metric}

All results for distance correlation~\cite{szekely2007measuring} of MJDs, including $p_{\text{norm}}$ and $p_{\text{sfmax}}$ with different $\tau$, are listed in Table~\ref{tab:correlation} for further exploring. There are many methods to achieve smoothness, and finding the most suitable one for the task at hand goes beyond the current focus of this paper. We will delve deeper into how smoothness can enhance performance in future work. 

\begin{table*}[htbp]
\centering
\resizebox{\textwidth}{!}{
\begin{tabular}{lcccc}
\toprule
\multicolumn{1}{c}{\multirow{2}{*}{\textbf{Distributions}}} & \multicolumn{1}{c}{$p_{\text{norm}}$} & \multicolumn{1}{c}{$p_{\text{sfmax}}$ $\tau=20$} & \multicolumn{1}{c}{$p_{\text{sfmax}}$ $\tau=10$} & \multicolumn{1}{c}{$p_{\text{sfmax}}$ $\tau=5$}\\ \cmidrule(lr){2-5} 
\multicolumn{1}{c}{}       
&  \textbf{\textbf{D.Corr  $\uparrow$}}&  \textbf{\textbf{D.Corr  $\uparrow$}}& \textbf{\textbf{D.Corr  $\uparrow$}}&  \textbf{\textbf{D.Corr  $\uparrow$}}    \\
\midrule
Chaos NLI  & 1 & -  &-&-\\
MNLI single label                                                                                   &0.612                       & -    & - & -          \\
MNLI distribution                                                                                    &0.795          & -    & - & -     \\
VariErr NLI                                                                                &0.688           & -    & - & -    \\
\midrule
Mixtral original                                                                   &0.609  &0.593  &0.584   &0.577                 \\
 4 explanations at a time
 &0.642&0.621 &0.612  &0.605\\
3  explanations at a time
 &0.659&0.636 &0.625 &0.617\\
 2  explanations at a time
&0.685  &0.662 &0.651&0.644\\
 1 explanations      at a time                                        &0.731  &0.713 &0.704  &0.697\\
\midrule
4  explicit explanations   at a time                                  &0.623 &0.608 &0.601  &0.596\\
3 explicit explanations    at a time                                &0.651  &0.637 &0.629 &0.624\\
2  explicit explanations   at a time                                 &0.678 &0.664  &0.606 &0.652\\
 1  explicit explanation at a time &0.719&0.709  &0.704 &0.701\\
\midrule
\midrule
Llama3 original                                                                  &0.689 &0.677    &0.665 &0.656    \\
 4  explanations    at a time                                           &0.750 &0.740  &0.730&0.714\\
3 explanations     at a time                                         &0.770   &0.763 &0.754&0.743\\
2 explanations    at a time                                          &0.795    &0.790 &0.783&0.774\\
1 explanations     at a time                                          &0.818 &0.812 &0.807 &0.797\\
\midrule
4  explicit explanations     at a time                                &0.733  &0.725 &0.716&0.703\\
 3 explicit explanations   at a time                                  &0.757  &0.752  &0.746&0.739\\
 2 explicit explanations at a time
&0.784 &0.779 &0.774&0.769\\
1  explicit explanation at a time
&0.809  &0.802 &0.796&0.787\\
\bottomrule
\end{tabular}}
\caption{Distance Correlation (D.Corr) results. All ablation results for comments numbers. Scores are compared with Chaos NLI's HJD.}\label{tab:correlation}
\end{table*}

\section{``Assistant'' Mode for Mixtral Prompt}
\label{app:assistant}

\begin{table*}[htbp]
\centering
\resizebox{\textwidth}{!}{
\begin{tabular}{lcccccc}
\toprule
\multicolumn{1}{c}{\multirow{2}{*}{\textbf{Distributions}}} & \multicolumn{3}{c}{$p_{\text{norm}}$} & \multicolumn{3}{c}{$p_{\text{sfmax}}$}\\ \cmidrule(lr){2-7} 
\multicolumn{1}{c}{}                                               & \textbf{\textbf{KL $\downarrow$}} & \textbf{\textbf{JSD $\downarrow$}} & \textbf{\textbf{TVD $\downarrow$}}  &  \textbf{\textbf{KL $\downarrow$}} & \textbf{\textbf{JSD $\downarrow$}} & \textbf{\textbf{TVD $\downarrow$}}      \\
\midrule
MNLI ground truth ~\cite{DBLP:conf/naacl/WilliamsNB18}                                                                               & 9.288          & 0.422  &0.435                 & -          & -                 & -             \\
MNLI distribution ~\cite{DBLP:conf/naacl/WilliamsNB18}                                                                               & 1.242          & 0.281   &0.295               & - &-  & -        \\
VariErr NLI ~\cite{DBLP:journals/corr/abs-2403-01931}                                                                            & 3.604           & 0.282   &0.296                & -          & -  & -       \\
Mixtral user best
 & \textbf{0.245}          & \textbf{0.211}    &\textbf{0.239}               & \textbf{0.217}          & \textbf{0.208}    &\textbf{0.232}            \\
Llama3 user best
             & \textbf{0.235}          & \textbf{0.247}   &\textbf{0.266}            & \textbf{0.212}          & \textbf{0.232}   &\textbf{0.245}           \\
\midrule
Mixtral original                                                                  & 0.433        & 0.291   & 0.340              & 0.434          & 0.292     &0.342                         \\
 4 explanations at a time
&0.266 &0.242 &0.270 &0.264 &0.241 &0.269  \\
3  explanations at a time
&0.260 &0.240 &0.268 &0.257 &0.239 &0.266   \\
 2  explanations at a time
&0.258 &0.239 &0.267 &0.252 &0.238 &0.265   \\
 1 explanations      at a time                                        &0.250 &0.237 &0.264 &0.245 &0.236 &0.262    \\
\midrule
4  explicit explanations   at a time                                  &0.224 &0.221 &0.243 &0.230 &0.223 &0.245    \\
3 explicit explanations    at a time                                 &0.220 &0.220 &0.242 &0.224 &0.221 &0.243    \\
2  explicit explanations   at a time                                  &0.216 &0.217 &0.239 &0.217 &0.218 &0.240    \\
 1  explicit explanations at a time&0.203 &0.213 &0.231 &0.198 &0.208 &0.227   \\
\bottomrule
\end{tabular}}
\caption{``Assistant'' mode results. Scores are compared with Chaos NLI's label probability distributions.}\label{tab:assistant}
\end{table*}

\begin{table}[htbp]
\scriptsize
\centering
\begin{tabular}{P{0.09\textwidth} | P{0.33\textwidth}}
\toprule 
\multicolumn{1}{l|}{\textbf{Type}} & \multicolumn{1}{l}{\textbf{General Instruction Prompt}} \\
\midrule
\texttt{With \newline explanations} & \textbf{"role": "user", "content"}: \newline Please add some comments for the relationship between Context and Statement. \newline Context: \{promise\} \newline Statement: \{hypothesis\} 
\newline \textbf{"role": "assistant", "content"}: \newline Comment 1: \{explanation 1\} \newline Comment 2: \{explanation 2\} \newline ...
\newline \textbf{"role": "user", "content"}: \newline Please carefully and fairly base your selection on the Comment to determine whether the Statement is true (Entailment), undetermined (Neutral), or false (Contradiction) given the Context and select ONE of the listed options and start your answer with a single letter.  \newline A. Entailment \newline B. Neutral \newline C. Contradiction. \newline Answer: \\
\midrule
\texttt{With explicit explanations} & \textbf{"role": "user", "content"}: \newline Please add some comments for the relationship between Context and Statement. \newline Context: \{promise\} \newline Statement: \{hypothesis\} 
\newline \textbf{"role": "assistant", "content"}: \newline Comment 1: \{explanation 1\}, so I choose {label1} \newline Comment 2: \{explanation 2\}, so I choose {label2} \newline ...
\newline \textbf{"role": "user", "content"}: \newline Please carefully and fairly base your selection on the Comment to determine whether the Statement is true (Entailment), undetermined (Neutral), or false (Contradiction) given the Context and select ONE of the listed options and start your answer with a single letter.  \newline A. Entailment \newline B. Neutral \newline C. Contradiction. \newline Answer: \\
\bottomrule
\end{tabular}

\caption{Instruction prompt for ``assistant'' mode.}
\label{tab:assis-prompts}
\end{table}

Here we'd introduce a really interesting way to prompt. Because this approach lacks a complete theoretical basis and currently proves effective only in Mixtral, not in Llama3, we are presenting the concept here without providing complete experimental evidence. We have proposed a special prompt method for chat templates based on Mixtral, which we refer to as ``assistant'' mode. We hope that anyone reading this who is interested or knowledgeable about this prompt method will be encouraged to further explore this approach.

We expected that LLM could focus more on the explanations in the ``assistant'' mode, since it is the text form that contains rich human label variation information. So we put the explanations in the content of \texttt{assistant} role as ``comments'' as shown in Table~\ref{tab:assis-prompts}. 
In that case, we divide the MCQA into two part, first to ask LLM about their opinion on the original NLI instances, then out human explanations in the position of LLM response (``assistant'' role).
LLM would assume that the comments are its own previous answers and take more account on them. 
And finally let LLM give a choice among three NLI labels.
It's a way to extract the potential of LLM's understanding capability for human label variation. Results are shown in Table~\ref{tab:assistant}.

We observed that in this approach, LLM's output becomes more sensitive to the input of explanations and seems to be more influenced by their content. However, our understanding is speculative. For example, Mixtral may emphasize ``assistant'' training during instruct tuning, or the architecture of a Mixture of Experts model might prioritize ``assistant'' content (as valuable global information to a specific expert). Yet, we lack the means to prove these speculations. Therefore, we present this as an interesting attempt in the appendix, acknowledging it as an exploration of prompt methods.

\begin{table}[htbp]
\scriptsize
\centering
\begin{tabular}{P{0.07\textwidth} | P{0.35\textwidth}}
\toprule 
\multicolumn{1}{l|}{\textbf{Role}} & \multicolumn{1}{l}{\textbf{Content}} \\
\midrule
\multicolumn{2}{l}{\textit{Instance from MNLI}}  \\
\midrule
Premise    & Krugman's column will henceforth be known as The Dismal Science, a phrase too famous to be ownable by anyone, except possibly British essayist Thomas Carlyle (1795-1881), who coined it.                         \\
\midrule
Hypothesis & Krugman writes novels.                      \\
\midrule
\multicolumn{2}{l}{\textit{Explanations from VariErr NLI}}  \\
\midrule
C0         & Krugman is a columnist, he doesn't write novels.                                                                                                                                          \\
\midrule
C1         & No, in the context is his column, which appears often in newspaper.                                                                                                                       \\
\midrule
N0         & It's clear that Krugman writes a column, but not clear whether they write novels.                                                                                                         \\
\midrule
N1         & Krugman has a column known as "The Dismal Science", it can be a novel column, but also can be others like essay column.                                  \\
\bottomrule
\end{tabular}
\caption{The sample for the case study. ``C0'', ``C1'' support Contradiction while ``N0'',``N1'' support Neutral.}
\label{tab:case_study}
\end{table}

\begin{figure}[htbp]

        \centering
        \includegraphics[width=\linewidth]{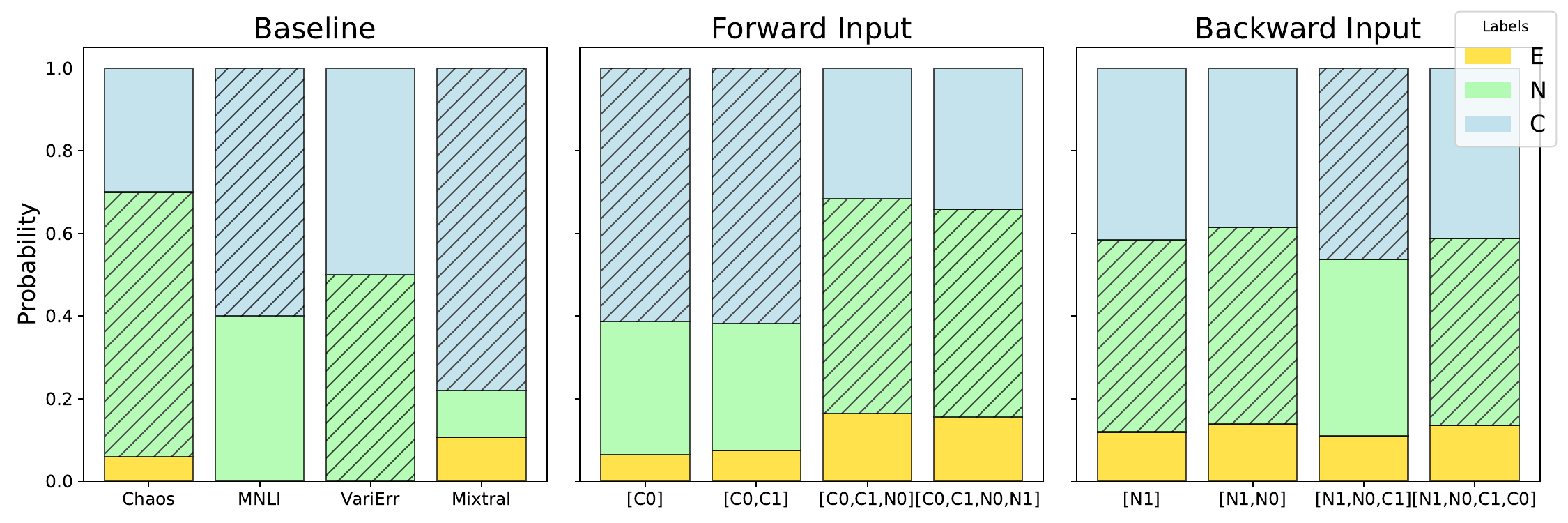}
        \caption{Results of case study. The three colored segments of each bar in the bar chart are labeled with the probabilities of three tags. The ticks below each chart represent the sequence combinations of the input explanations.``C0'', ``C1'' are explanations supporting Contradiction while ``N0'',``N1'' support Neutral. Detailed of the used sample are in Table~\ref{tab:case_study}.}
        \label{fig:case_study}
    
\end{figure}

\section{Case Study Details}
\label{app:case_study}

We conducted a case study to demonstrate how explanations influence LLM's output, and empirically showed that the order of explanations does indeed impact the model's output. This led us to propose the approach in \S\ref{subsec:estimate-bias}, which involves reducing the bias introduced by different sequences through permutation and averaging.

The sample used is shown in Table~\ref{tab:case_study} and the results are shown in Figure~\ref{fig:case_study}. More details of the output logits and texts were recorded in Table~\ref{tab:app_case_study}. We use Mixtral with ``assistant'' mode to implement this case study.

Through the results, we observed that the output label probability of Mixtral changes, when increasing the number of explanations, whether in a forward direction or backward direction. And, the order does impact. Latter ones usually impact more, but the final results could be similar, which means LLM can capture the HLV information inside, even when facing multiple explanations with diverse opinions.

\begin{table*}[t]
\scriptsize
\centering
\resizebox{\textwidth}{!}{
\begin{tabular}{P{0.14\linewidth} | P{0.81\linewidth}}
\toprule 
\multicolumn{1}{l|}{\textbf{Setting}} & \multicolumn{1}{l}{\textbf{Content}} \\
\midrule
no explanations & - \\
scores/probs & [5.906385898590088, 6.259021282196045, 43.25299835205078] / [0.106578055463, 0.1129412010, 0.78048074346] \\
output text & C. Contradiction. 

The context provided does not give any information about Krugman writing novels. Instead, it mentions that Krugman's column will be named "The Dismal Science," a phrase coined by Thomas Carlyle. Therefore, the statement that Krugman writes novels is a contradiction given the context.</s> \\
\midrule
\multicolumn{2}{l}{\textit{Foward Input}}   \\
\midrule
input explanations [C0] & Comment 1: Krugman is a columnist, he doesn't write novels.    \\
scores/probs  & [4.2198514938, 20.7870941162, 39.63526535] / [0.065280123175, 0.32157152126, 0.61314835556] \\
output text & C. Contradiction. The statement "Krugman writes novels" is a contradiction given the context "Krugman's column will henceforth be known as The Dismal Science" as it implies that Krugman is a columnist and not a novelist.</s>\\
\midrule
input explanations [C0,C1] & Comment 1: Krugman is a columnist, he doesn't write novels.

Comment 2: No, in the context is his column, which appears often in newspaper\\
scores/probs  & [4.7512874603, 19.5720710754, 39.330707550] / [0.074642324559, 0.30747558292, 0.61788209251] \\
output text & C. Contradiction.

Explanation: The context states that Krugman writes a column called "The Dismal Science," and that the phrase was coined by Thomas Carlyle. The statement, "Krugman writes novels," contradicts this context, as it suggests that Krugman writes novels, rather than a column.</s>\\
\midrule
input explanations [C0,C1,N0] &Comment 1: Krugman is a columnist, he doesn't write novels. 

Comment 2: No, in the context is his column, which appears often in newspaper 

Comment 3: It's clear that Krugman writes a column, but not clear whether they write novels. \\
scores/probs  & [11.1629428863, 35.1051597595, 21.4133796691]/ [0.16493348704, 0.51868189878, 0.31638461417] \\
output text & B. Neutral

The context only provides information about Krugman's column being known as "The Dismal Science" and that the phrase was coined by Thomas Carlyle. It does not provide any information about Krugman writing novels. Therefore, the statement "Krugman writes novels" is neutral, neither entailed nor contradicted by the context.</s>\\
\midrule
input explanations [C0,C1,N0,N1] & Comment 1: Krugman is a columnist, he doesn't write novels. 

Comment 2: No, in the context is his column, which appears often in newspaper 

Comment 3: It's clear that Krugman writes a column, but not clear whether they write novels. 

Comment 4: Krugman has a column known as "The Dismal Science", it can be a novel column, but also can be others like essay column. \\
scores/probs & [10.3001527786, 33.403881072, 22.699693679] / [0.15511407509, 0.5030422585, 0.34184366636] \\
output text &B. Neutral

The statement "Krugman writes novels" is not directly addressed or contradicted by the context provided. The context only mentions that Krugman's column will be known as "The Dismal Science" and that the phrase was coined by Thomas Carlyle. It does not provide any information about Krugman's novel-writing activities. Therefore, the statement is undetermined based on the context.</s> \\
\midrule
\multicolumn{2}{l}{\textit{Backward Input}}   \\
\midrule
input explanations [N1] &Comment 1: Krugman has a column known as "The Dismal Science", it can be a novel column, but also can be others like essay column.  \\
scores/probs & [8.8821163177, 34.580959320, 30.8163871765] / [0.119577013365, 0.46555209219, 0.414870894442] \\
output text &B. Neutral. The context only provides information about the name of Krugman's column, "The Dismal Science", and a brief history of the phrase. It does not provide any information about Krugman writing novels, so the statement "Krugman writes novels" cannot be determined as true or false based on the given context.</s> \\
\midrule
input explanations [N1,N0] & Comment 1: Krugman has a column known as "The Dismal Science", it can be a novel column, but also can be others like essay column. 

Comment 2: It's clear that Krugman writes a column, but not clear whether they write novels.\\
scores/probs &[9.7944955825, 33.278915405, 26.9777603149] / [0.139819155060, 0.475065795280, 0.385115049658] \\
output text & B. Neutral. The context does not provide enough information to confirm or deny whether Krugman writes novels. The column being referred to as "The Dismal Science" does not have any impact on Krugman's novel-writing abilities or activities.</s>\\
\midrule
input explanations [N1,N0,C1] &Comment 1: Krugman has a column known as "The Dismal Science", it can be a novel column, but also can be others like essay column. 

Comment 2: It's clear that Krugman writes a column, but not clear whether they write novels. 

Comment 3: No, in the context is his column, which appears often in newspaper \\
scores/probs & [8.103030204, 31.667663574, 34.176952362] / [0.109577932870, 0.42824437594, 0.462177691185] \\
output text & C. Contradiction.

Explanation: The context states that Krugman's column will be known as "The Dismal Science", and it does not provide any information about Krugman writing novels. Therefore, the statement "Krugman writes novels" is not true based on the given context.</s>\\
\midrule
input explanations [N1,N0,C1,C0] &Comment 1: Krugman has a column known as "The Dismal Science", it can be a novel column, but also can be others like essay column. 

Comment 2: It's clear that Krugman writes a column, but not clear whether they write novels. 

Comment 3: No, in the context is his column, which appears often in newspaper 

Comment 4: Krugman is a columnist, he doesn't write novels. \\
scores/probs & [9.7341032028, 32.308975219, 29.411777496] / [0.136227315521, 0.452159266214, 0.411613418264] \\
output text & B. Neutral. The context does not provide enough information to confirm or deny whether Krugman writes novels.</s>\\
\bottomrule
\end{tabular}}
\caption{Case study records.}
\label{tab:app_case_study}
\end{table*}